\documentclass[lettersize,journal]{IEEEtran}

\usepackage[numbers]{natbib}

\ifCLASSINFOpdf
  \usepackage[pdftex]{graphicx}
\else
\fi

\usepackage{amsmath}

\usepackage{algorithmic}

\usepackage{array}

\ifCLASSOPTIONcompsoc
 \usepackage[caption=false,font=footnotesize,labelfont=sf,textfont=sf]{subfig}
\else
 \usepackage[caption=false,font=footnotesize]{subfig}
\fi

\usepackage{url}

\usepackage{booktabs}
\usepackage{bbm}
\usepackage{amsmath}
\usepackage{amssymb}
\usepackage{amsthm}
\usepackage{multirow}
\usepackage{tcolorbox}
\pdfoutput=1

\definecolor{DarkGreen}{RGB}{38,135,94}
\definecolor{LightGreen}{RGB}{238,247,243}
\newtcolorbox{greenbox}[1]{colback=LightGreen!35!white,
    colframe=DarkGreen!75!black,fonttitle=\bfseries,
    title={#1}}

\definecolor{DarkRed}{RGB}{184,31,0}
\definecolor{LightRed}{RGB}{218,118,91}
\newtcolorbox{redbox}[1]{colback=LightRed!15!white,
    colframe=DarkRed!75!black,fonttitle=\bfseries,
    title={#1}}

\newcommand{\augsimclr}{\mathcal{T}_\text{SimCLR}}

\newcommand{\augclass}{\mathcal{T}_\text{class}}

\newcommand{\misimclr}{\hat{I}_\text{SimCLR}(h_X;h_Y)}
\newcommand{\miclass}{\hat{I}_\text{class}(h_X;h_Y)}

\newtheorem{proposition}{Proposition}[section]
\newtheorem{theorem}{Theorem}[section]

\newtheorem{definition}{Definition}[section]

\begin{document}

\title{Towards a Rigorous Analysis of \\ Mutual Information in Contrastive Learning}

\author{Kyungeun~Lee, Jaeill~Kim, Suhyun~Kang, and~Wonjong~Rhee,~\IEEEmembership{Fellow,~IEEE}
\IEEEcompsocitemizethanks{\IEEEcompsocthanksitem The authors are with Deep Representation Learning Research Group, Department of Intelligence and Information, Seoul National University, Seoul, South Korea. 
E-mail: \{ruddms0415,~jaeill0704,~su\_hyun,~wrhee\} @snu.ac.kr
\IEEEcompsocthanksitem Kyungeun Lee is currently with LG AI research, Seoul, South Korea. This work was conducted while at Seoul National University.
\IEEEcompsocthanksitem Wonjong Rhee is also with IPAI (Interdisciplinary Program in Artificial Intelligence).}
\thanks{Manuscript received ...}}

\markboth{Journal of \LaTeX\ Class Files,~Vol.~14, No.~8, August~2015}%
{Shell \MakeLowercase{\textit{et al.}}: Bare Demo of IEEEtran.cls for Computer Society Journals}

\IEEEtitleabstractindextext{%
\begin{abstract}
Contrastive learning has emerged as a cornerstone in recent achievements of unsupervised representation learning. Its primary paradigm involves an instance discrimination task with a mutual information loss. The loss is known as InfoNCE and it has yielded vital insights into contrastive learning through the lens of mutual information analysis.
However, the estimation of mutual information can prove challenging, creating a gap between the elegance of its mathematical foundation and the complexity of its estimation. As a result, drawing \textit{rigorous} insights or conclusions from mutual information analysis becomes intricate.
In this study, we introduce three novel methods and a few related theorems, aimed at enhancing the rigor of mutual information analysis. Despite their simplicity, these methods can carry substantial utility.
Leveraging these approaches, we reassess three instances of contrastive learning analysis, illustrating their capacity to facilitate deeper comprehension or to rectify pre-existing misconceptions. Specifically, we investigate small batch size, mutual information as a measure, and the InfoMin principle.

\end{abstract}

}

\maketitle

\IEEEdisplaynontitleabstractindextext

\IEEEpeerreviewmaketitle

\vspace{1cm}

\vspace{-1.0cm}

\section{Introduction}
\label{sec:intro}


Contrastive learning~\cite{oord2018cpc,chen2020simple} has achieved a remarkable success in the field of unsupervised representation learning. 
The key elements driving the success of contrastive learning have been recognized to be instance discrimination and InfoNCE loss.
Instance discrimination~\cite{wu2018unsupervised,wu2018npid} 
focuses on learning a robust feature representation that captures inherent similarities among the augmented instances generated from a single input.  
\textit{InfoNCE loss}~\cite{oord2018cpc,chen2020simple,poole2019variational} serves as the training objective. 
It not only assumes a central role in attaining outstanding performance, but it also provides an elegant framework for interpreting contrastive learning as the maximization of \textit{Mutual Information} (MI) between the two augmented views, denoted as $X$ and $Y$.
Numerous works have studied contrastive learning based on the MI interpretation, with some becoming pivotal for comprehending the nuances of contrastive learning~\cite{oord2018cpc,hjelm2018learning,tian2020contrastive,tian2020makes}. 
However, conducting an investigation based on the mutual information of $X$ and $Y$ can be difficult, potentially leading to an incorrect analysis.


Mutual Information (MI) has played a critical role in improving and understanding deep networks. 
Formally, the MI between $X$ and $Y$, with joint distribution $p(x,y)$ and marginal distributions $p(x)$ and $p(y)$, is defined as
\begin{equation}
    I(X;Y)=\mathbb{E}\left[ \log{\frac{p(x,y)}{p(x)p(y)}} \right].
    \label{eq:1}
\end{equation}
MI quantifies the shared Shannon information between two random variables and can serve as a fundamental measure of dependency~\cite{cover1999elements}. 
Several intrinsic challenges arise in the mutual information analysis of deep learning models, with five key factors outlined below.
\begin{enumerate}
    \item \textit{The joint distribution $p(x,y)$ is not accessible.} Therefore, an exact evaluation of MI is not possible, necessitating the utilization of estimation methods. 

    \item \textit{The dimensions of $X$ and $Y$ are substantially high in the context of practical deep learning problems.} As a consequence, conventional estimators~\cite{kraskov2004estimating,gao2015efficient}
    become ineffective, prompting the adoption of neural estimators~\cite{poole2019variational,belghazi2018mine,song2019understanding}.
    Neural estimators rely on variational bounds parameterized by neural networks.

    \item \textit{The accuracy and reliability of the latest neural estimators have not been thoroughly understood.}     
    Existing investigations have predominantly focused on assessing their reliability within Gaussian distributions, where analytical evaluations are feasible~\cite{poole2019variational,belghazi2018mine,song2019understanding}. However, in the case of 
    practical datasets like images, their accuracy remains insufficiently characterized.

    \item \textit{Drawing rigorous interpretations based on bounds can be tricky.} 
    Mainstream neural estimators commonly depend on utilizing either a lower bound or an upper bound of MI, along with its analytical reformulation, to facilitate the variational calculations. Consequently, the estimated value must be regarded as a bound value for a rigorous interpretation. This introduces complexities in deriving meaningful insights. In the existing works, the estimated value is frequently treated as the true value for the purpose of interpretation.

    \item \textit{Identifying the specific shared information contributing to MI can be challenging.} MI serves as a numerical gauge of shared information, devoid of revealing the distinct contributions from individual information components. Therefore, unraveling the degree to which specific pieces of information contribute becomes intricate, potentially introducing complexities in the MI analysis. 

\end{enumerate}

As we will demonstrate through three case studies detailed in Section~\ref{sec:results}, it becomes imperative to meticulously address the aforementioned challenges to prevent misconceptions arising from MI analysis.
Nonetheless, an earnest endeavor to enhance analysis rigor can potentially lead to a drawback, leading to inconclusive outcomes for a substantial portion of MI analyses.
To overcome this drawback, we propose three effective methods to bolster MI analysis.
The first method is \textit{same-class-sampling}. In contrastive learning, the choice of augmentation dictates the shared information between the two views, where the choice directly commands the joint distribution $p(x,y)$, in turn $p(x,y)$ decides the MI of learning, and ultimately the MI affects what will be learned as the representation. The same-class sampling shares only the class information between the two views and its true MI can be proven to be the same as the class entropy $H(C)$ when the estimate is the same as $H(C)$.
The second method involves the \textit{utilization of the CDP dataset}. 
This dataset enables embodiment of information into an image through manipulation of color, digit, and position.
Notably, the true MI value can be readily controlled by adjusting the dependency between the two views.
Our main utilization of the CDP dataset is in crafting experiments that allow concrete interpretations of the outcomes.
The third method is \textit{post-training MI estimation}. In the previous works, MI estimation was commonly performed during the training process because InfoNCE can serve both as a training loss and a variational estimator for MI. However, decoupling the MI estimation and moving it to a post-training phase offers a few advantages. In particular, this separation allows for a comprehensive comparison of diverse representation encoders because the post-training MI estimation can be applied to any pre-trained network regardless of its supervised or unsupervised training method.

To demonstrate how we can make a progress toward a more rigorous MI analysis, we present three case studies where we re-examine previously established insights. The case studies extensively employ the three proposed methods, with the established insights outlined as follows:

\begin{enumerate}[leftmargin=20pt]
    \item A small batch size is undesirable for contrastive learning because of InfoNCE's $\mathcal{O}(\log{K})$ bound~\cite{chen2020simple,tian2020contrastive,hjelm2018infomax,bachman2019learning,sordoni2021decomposed,wu2020conditional,song2020mlcpc}.
    \item  Large MI is not predictive of downstream performance~\cite[\S3.1]{tschannen2019mutual}.
    Instead, other metrics such as uniformity~\cite{wang2020understanding,wang2021understanding}, alignment~\cite{wang2020understanding}, and tolerance~\cite{wang2021understanding}
    are more relevant and useful.
    \item In contrastive learning, task-irrelevant information needs to be discarded for a better generalization~\cite{tian2020makes, wang2022rethinking, tsai2020self, xiao2020should, chen2021intriguing}.
\end{enumerate}


The rest of this paper is organized as following. 
Section~\ref{sec:infonce} provides an overview of the InfoNCE loss and the InfoNCE estimation of MI.
Section~\ref{sec:method} elaborates on the three proposed methods in detail.
Section~\ref{sec:results} presents the three case studies in depth. 
Section~\ref{sec:discussions} provides discussions on two subjects. First, we discuss MI's limitation as the learning objective despite its excellence as a metric of representation quality. Second, we discuss strategies that surpass contrastive learning, driving the progress of unsupervised learning even further.
Section~\ref{sec:Conclusion} contains the conclusions.

\section{InfoNCE for learning and estimation} 
\label{sec:infonce}
In the absence of annotation information, contrastive learning generates multiple views of a given image through augmentations and learns useful representations by pursuing an invariance over the multiple views. This has been known as the \textit{instance discrimination}~\cite{wu2018npid}, and it has been empirically found that \textit{InfoNCE loss}~\cite{oord2018cpc,gutmann2010noise} is an effective training objective for a variety of downstream tasks. Given an unannotated dataset $\mathcal{D} = \left\{ s_i | s_i \in \mathbb{R}^{m} \right\}$, we can sample an image $s_i$ and randomly generate its first and second views with a family of augmentations $\mathcal{T}$ to form a positive pair $(x_i, y_i)$. See Figure~\ref{fig:fig1}(b) for an example where $\mathcal{T}$ is a family of SimCLR~\cite{chen2020simple} augmentations. After repeating it $K$ times to sample $K$ pairs, InfoNCE loss for the batch can be calculated as the following. 
\begin{definition}[InfoNCE Loss \cite{oord2018cpc,chen2020simple}]
    \label{definition:infonce_estimation}
    The InfoNCE loss for contrastive learning is defined as 
    \begin{equation}
        \label{eq:infoNCE_loss}
        \mathcal{L}_\textit{InfoNCE} = \frac{1}{2K} \sum_{k=1}^{K} [l(x_k,y_k)+l(y_k,x_k)] 
    \end{equation}
    where $l(x_i,y_i)$ is given as 
    \begin{equation}
        l(x_i,y_i)=-\log{\frac{e^{ z_{x_i,y_i} / \tau }}{\sum_{j=1}^{K}\mathbbm{1}_{[j\neq i]}{e^{ z_{x_i,x_j} / \tau }} + \sum_{j=1}^{K}{e^{ z_{x_i,y_j}/\tau }}}}
    \end{equation}
    with $z_{x,y}=\text{sim}(f(x),f(y))$; $f=f_p\circ f_e$ with $f_e(\cdot)$ as the encoder and $f_p(\cdot)$ as the projection head; $\text{sim}(u, v) = u^Tv/||u|| ||v||$ denotes the dot product between $l_2$ normalized $u$ and $v$ (i.e. cosine similarity); $\tau$ denotes a temperature scalar; and $K$ is the batch size. We denote the encoded representation vector of an input $X$ as $h_X=f_e(X)$.
\end{definition}
The specific implementation of InfoNCE loss in Eq.~\eqref{eq:infoNCE_loss} is commonly addressed as NT-Xent~(normalized temperature-scaled cross-entropy) loss, following the naming in~\cite{chen2020simple}.
%
While the InfoNCE loss is for learning, a variational bound based on InfoNCE can be used for estimating mutual information. The estimated MI can be defined as the following.
\begin{definition}[Estimated MI \cite{oord2018cpc,poole2019variational}]
    \label{definition:estimated_mi}
    The InfoNCE estimation of mutual information is defined as 
    \begin{equation}
    \hat{I}(X;Y) \triangleq \log{(2K-1)} - \mathcal{L}_\textit{InfoNCE}. \label{eq:definition:estimated_mi}
    \end{equation}
\end{definition}
The intimate relationship between the InfoNCE loss and the estimated MI has been studied before. For instance, see \cite{oord2018cpc,poole2019variational} for a derivation.
In our study, we provide the following two previously known propositions and their proofs for the completeness. 

\begin{proposition}[InfoNCE estimation as a lower bound of the true MI \cite{oord2018cpc,poole2019variational}]
    \label{proposition:true_estimated_MI}
    The InfoNCE estimation of mutual information is a lower bound of the true mutual information.
    \begin{equation}
    \hat{I}(X;Y) \le I(X;Y)  \label{eq:proposition:true_estimated_MI}
    \end{equation}
\end{proposition}

\begin{proposition}[$\log{(2K-1)}$ Bound \cite{oord2018cpc,poole2019variational}]
    \label{proposition:infonce_2Kminus1_inequality}
    The InfoNCE estimation of mutual information is upper bounded by $log{(2K-1)}$. 
    \begin{equation}
    \hat{I}(X;Y) \le \log{(2K-1)} \label{eq:proposition:infonce_2Kminus1_inequality}
    \end{equation}
\end{proposition}
\begin{proof}
    The proof is based on the variational bound derivation. Let $q(x|y)=\frac{p(x)}{Z(y)}e^{z_{x,y}/\tau},$ where $Z(y)=\mathbb{E}_{p(y)}[e^{z_{x,y}/\tau}]$. Then the true MI, $I(X;Y)$, can be bounded as the following. 
    \begin{align}
        &I(X;Y) \nonumber \\
        &= \mathbb{E}_{p(x,y)}\left[ \log{\frac{p(x,y)}{p(x)p(y)}} \right] = \mathbb{E}_{p(x,y)}\left[\log{\frac{p(x|y)}{p(x)}}\right] \\
        &= \mathbb{E}_{p(x,y)}\left[\log{\frac{q(x|y)}{p(x)}}\right] + \mathbb{E}_{p(y)}[KL(p(x|y)||q(x|y))]\\
        &\ge \mathbb{E}_{p(x,y)}\left[\log{\frac{q(x|y)}{p(x)}}\right] \label{eq12}\\
        & =\mathbb{E}_{p(x,y)}\left[\log{\frac{e^{z_{x,y}/\tau}}{Z(y)}}\right] \\
        &\approx \mathbb{E}\left[\log{\frac{e^{z_{x_i,y_i}/\tau}}{\frac{1}{2K-1}
        \sum\limits_{j=1}^{K} \left(\mathbbm{1}_{[j\neq i]}{e^{ z_{x_i,x_j} / \tau }} + e^{ z_{x_i,y_j}/\tau } \right)
        }}\right] \label{eq6}\\  
        &=\log{(2K-1)} \label{eq7} \\ &\phantom{abc}+ \mathbb{E}\left[\log{\frac{e^{z_{x_i,y_i}/\tau}}{
        \sum\limits_{j=1}^{K} \left(\mathbbm{1}_{[j\neq i]}{e^{ z_{x_i,x_j} / \tau }} + e^{ z_{x_i,y_j}/\tau } \right)
        }}\right] \label{eq8} \\
        &=\log{(2K-1)} - \mathcal{L}_\textit{InfoNCE} \label{eq9} \\ 
        &\triangleq \hat{I}(X;Y)
    \end{align}
    The inequality in Eq.~\eqref{eq12} is due to the non-negativeness of KL-divergence, and the approximation in Eq.~\eqref{eq6} is due to the replacement of the expectation with its empirical mean. 

    The proof of Proposition~\ref{proposition:true_estimated_MI} is directly obtained from the above derivation of InfoNCE variational bound. The proof of Proposition~\ref{proposition:infonce_2Kminus1_inequality} also follows from the derivation. In Eq.~\eqref{eq9}, the term $-\mathcal{L}_\textit{InfoNCE}$ is always negative because the argument of the $\log$ term in Eq.~\eqref{eq8} is always between zero and one. This can be easily confirmed because the denominator term $\sum\limits_{j=1}^{K} \left(\mathbbm{1}_{[j\neq i]}{e^{ z_{x_i,x_j} / \tau }} + e^{ z_{x_i,y_j}/\tau } \right)$ is a sum of positive values and because the summation includes the numerator term $e^{ z_{x_i,x_j} / \tau }$.
\end{proof}

Note that the bound in Proposition~\ref{proposition:infonce_2Kminus1_inequality} has been known as $\log(K)$ bound, but the actual bound for NT-Xent implementation is $\log{(2K-1)}$ because the number of denominator terms in Eq.~\eqref{eq6} is $2K-1$. 
Also, note that we can see from Eq.~\eqref{eq:definition:estimated_mi} that minimizing $\mathcal{L}_\textit{InfoNCE}$ is equivalent to maximizing MI estimation $\hat{I}(h_X;h_Y)$.

\section{Methods for MI analysis}
\label{sec:method}
%
In this section, we explain three methods that can be helpful for overcoming the challenges of MI analysis listed in Section~\ref{sec:intro}. They are same-class sampling, use of CDP dataset, and post-training MI estimation. The second one is actually an introduction of a simple dataset and its use as a method is explained. 


\subsection{Same-class sampling}
\label{subsec:same_class_sampling}

\begin{figure*}[t!]
    \centering
    \begin{minipage}{0.2\textwidth} 
    \subfloat[Positive pairing]{
    \includegraphics[width=0.9\textwidth]{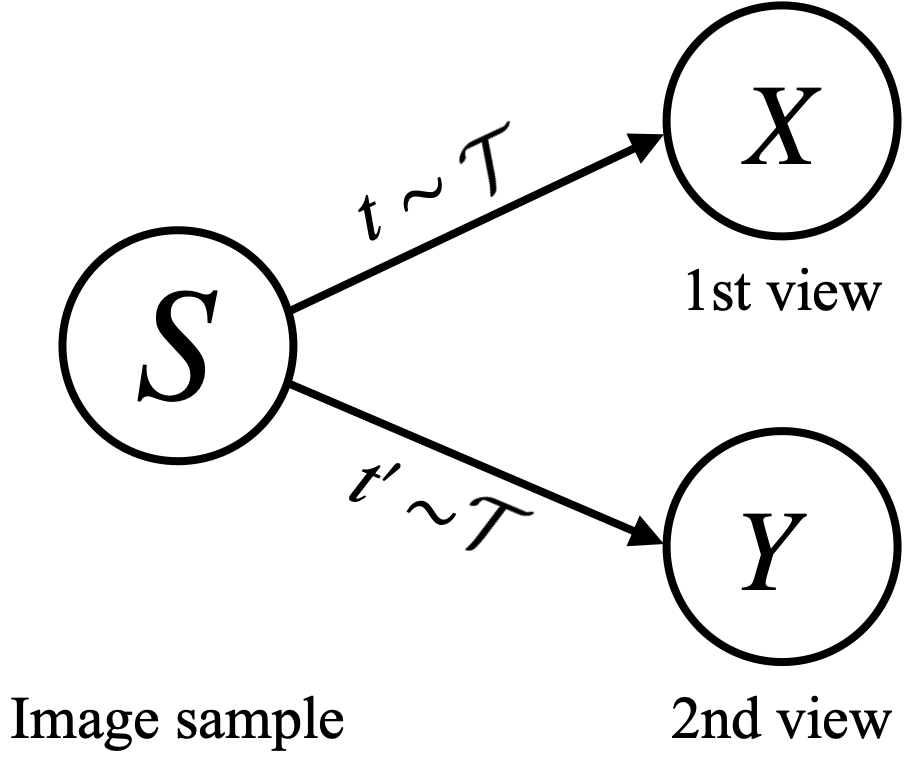}
    }
    \end{minipage} 
    \hspace{0.3cm}\vline\hspace{0.3cm}
    \begin{minipage}{0.7\textwidth}
    \subfloat[An example of positive pairing using a family of augmentations $\mathcal{T}_\text{aug}$. SimCLR augmentation~\cite{chen2020simple} is shown.
    ]{
    \includegraphics[width=\textwidth]{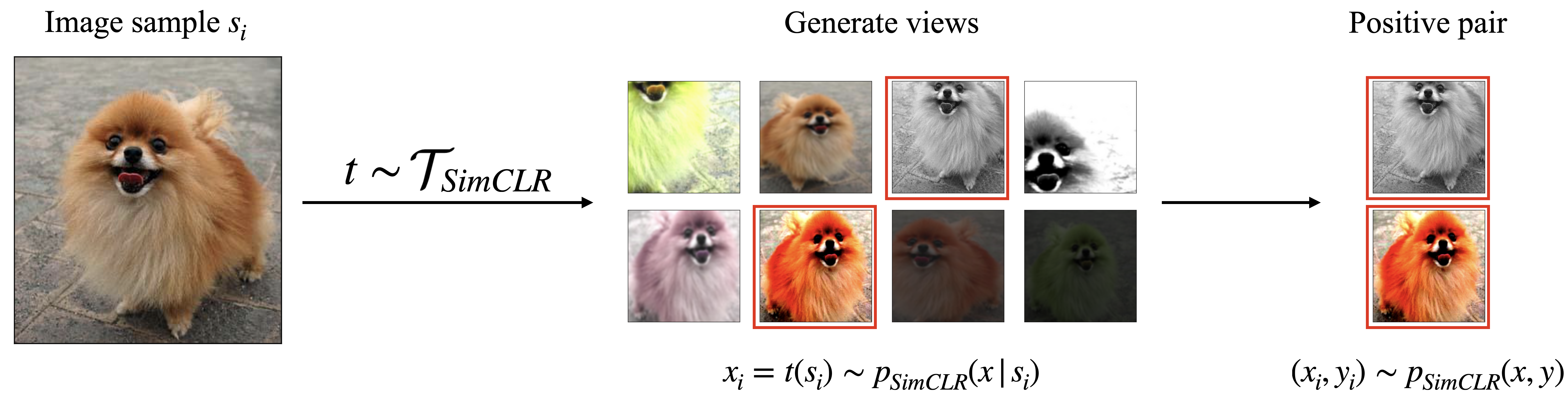}
    } \\
    \subfloat[Positive pairing with same-class sampling $\augclass$. Unlike the case of using a family of augmentations $\mathcal{T}_\text{aug}$, only downstream task's class information is used for positive pairing.
    ]{
    \includegraphics[width=\textwidth]{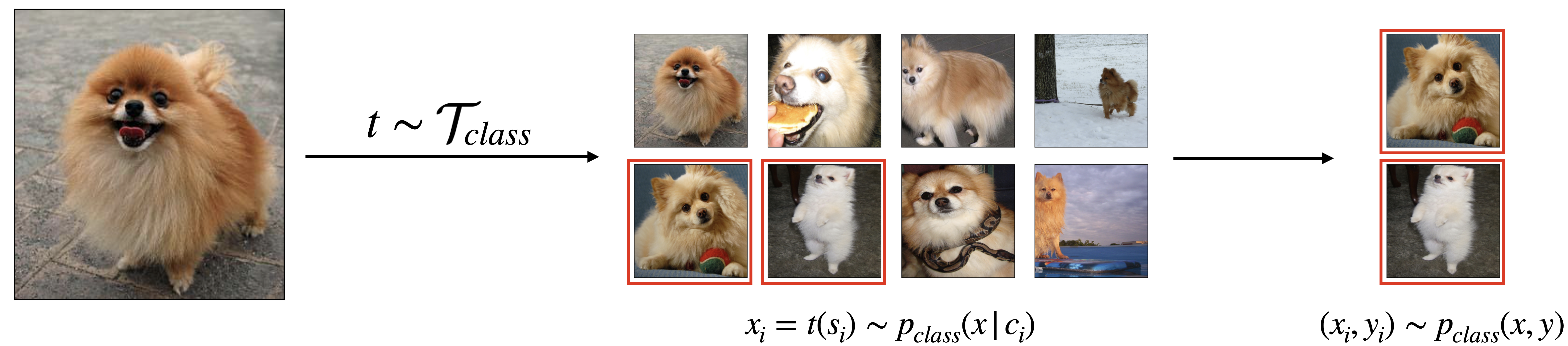}
    }
    \end{minipage}
    \caption{
    Positive pairing method implicitly determines the joint distribution: $p(x,y)$ is determined by the choice of $\mathcal{T}$.
    }
    \label{fig:fig1}
\end{figure*}

MI is a function of joint distribution $p(x,y)$ and marginal distribution $p(x)p(y)$ as defined in Eq.(\ref{eq:1}). In contrastive learning, the choice of augmentation $\mathcal{T_\text{aug}}$ directly determines the joint and marginal distributions of the paired views. Therefore, the design of $\mathcal{T_\text{aug}}$ plays the central role in contrastive learning where it regulates the learning objective through the InfoNCE loss. In many application fields, the main research work for contrastive learning is to design and tune $\mathcal{T_\text{aug}}$.

Despite the fundamental importance of the joint and marginal distributions, we do not need to know the exact distributions for performing training. Instead, we just need to be able to sample from a relevant and useful joint distribution for the given downstream task. While augmentation is a convenient way for enabling such a relevant and useful sampling, it is not a requirement as long as we can sample paired views that are meaningful. Therefore, the concept of augmentation $\mathcal{T}_\text{aug}$ can be expanded to the concept of positive pairing $\mathcal{T}$ as shown in Figure~\ref{fig:fig1}. The positive pairing in Figure~\ref{fig:fig1}(a) encompasses augmentation based (e.g., SimCLR augmentation in Figure~\ref{fig:fig1}(b)) and non-augmentation based sampling methods (e.g., same-class sampling in Figure~\ref{fig:fig1}(c)). In our work, the simple yet special positive pairing method that does not rely on augmentation in Figure~\ref{fig:fig1}(c) is named as \textit{same-class sampling} -- $\mathcal{T}_\text{class}$. The same-class sampling only relies on the downstream task's label information and it does not utilize any augmentation at all.

Before moving forward, we emphasize that the use of downstream task's label information makes the same-class sampling practically useless for the purpose of unsupervised representation learning. By the use of label information, same-class sampling becomes a supervised representation learning method. Despite the practical limitation, same-class sampling is definitely worth studying because it allows theoretical developments that lead to special cases where the true mutual information values can be easily evaluated. As will become obvious in Section~\ref{sec:results}, the accessibility to the true MI value can play a key role when performing an MI analysis.

Now, we move on to the theoretical developments for same-class sampling. For the same-class sampling, the true MI for its joint distribution $p_\text{class}(x,y)$ can be proven to be upper bounded by the entropy of the class distribution $H(C)$. The proof is straightforward, and we formally provide the proposition below.


\begin{proposition}
    \label{proposition:inequality}
    For the same-class sampling $\mathcal{T}_\text{class}$ with its joint distribution $p_\text{class}(x,y)$, the mutual information between the first view $X$ and the second view $Y$ is \textbf{upper bounded} by the class entropy.
    \begin{equation}
    I_\text{class}(X;Y)\le H(C)
    \end{equation}
\end{proposition}

\begin{proof}
From the construction of same-class sampling, the dependency can be expressed as $X\leftarrow C \rightarrow Y$. The dependency is Markov equivalent to $X\rightarrow C\rightarrow Y$ because both Markov chains encode the same set of conditional independencies. Then,
\begin{align}
    I(X;Y) &\le I(X;C) \\
           &= H(C) - H(C|X) \\
           &\le H(C),
\end{align}
where the first inequality follows from the data processing inequality~\cite{cover1999elements} and the second inequality follows from the entropy's positiveness for discrete random variables.
\end{proof}

With Proposition~\ref{proposition:true_estimated_MI} and Proposition~\ref{proposition:inequality}, the following main theorem can be obtained.
\begin{theorem}
    \label{theorem:equality}
    For the same-class sampling $\mathcal{T}_\text{class}$ with its joint distribution $p_\text{class}(x,y)$, the mutual information between the first view $X$ and the second view $Y$ is \textbf{equal to} the class entropy when the estimated mutual information is equal to the class entropy.  
    \begin{align}
        & \hat{I}_\text{class}(X;Y) = H(C) \\
        \Rightarrow \ \ & I_\text{class}(X;Y) = H(C) \label{eq:theorem:equality}
    \end{align}
\end{theorem}
\begin{proof}
The following is a direct result of Proposition~\ref{proposition:true_estimated_MI} and Proposition~\ref{proposition:inequality}.
    \begin{equation}
        \hat{I}_\text{class}(X;Y) \le I_\text{class}(X;Y) \le H(C)
    \end{equation}
Because the true MI $I_\text{class}(X;Y)$ is in the middle, $\hat{I}_\text{class}(X;Y) = H(C)$ means that all three are of the same value.  
\end{proof}
For the most popular downstream benchmarks, such as CIFAR and ImageNet,  the calculation of $H(C)$ is trivial. Now, thanks to the Theorem~\ref{theorem:equality}, we can identify the true MI with no ambiguity whenever $\hat{I}_\text{class}(X;Y) = H(C)$ is satisfied. This theorem will be utilized as a key enabler for a rigorous MI analysis in Section~\ref{sec:results}.

Theorem~\ref{theorem:equality} can be useful when the true MI value is required for an MI analysis. The theorem, however, is not useful when the condition $\hat{I}_\text{class}(X;Y) = H(C)$ is not satisfied. For such a case, we derive another equality that can be proven under an error-free classifier assumption. 
\begin{theorem}
    \label{theorem:equality2_errorfree}
    For the same-class sampling $\mathcal{T}_\text{class}$ with its joint distribution $p_\text{class}(x,y)$, the mutual information between the first view $X$ and the second view $Y$ is \textbf{equal to} the class entropy when there exists an error-free classification function $f_\text{class}(\cdot)$. 
    \begin{align}
        & \exists \ \text{An error-free classifier} \ f_\text{class}(\cdot):X\rightarrow C \\
        \Rightarrow \ \ & I_\text{class}(X;Y) = H(C) \label{eq:theorem:equality2_errorfree}
    \end{align}
\end{theorem}
In reality, such an error-free classification function $f_\text{class}$ does not exist for practical and interesting problems. Nonetheless, the equality result is useful for understanding MI in a high-accuracy regime. The proof is provided below.
\begin{proof}
    From the construction of same-class sampling, the dependency can be expressed as $S\rightarrow C \rightarrow X$ and $S\rightarrow C \rightarrow Y$ where $C$ is the class label of the sampled source image $S$. Because of the error-free classifier $f_\text{class}(\cdot)$, the class label information can be perfectly extracted from $X$ or $Y$. This means that $X \rightarrow C$ and $Y \rightarrow C$ also hold. Using the dependencies, we can conclude that the following is a valid Markov chain. 
    \begin{equation}
    S \rightarrow C \rightarrow X  \rightarrow C \rightarrow Y \rightarrow C    \label{eq:Markov_chain}
    \end{equation}
    The desired equality proof can be obtained by deriving an upper bound $I(X;Y) \le H(C)$ and a lower bound $H(C) \le I(X;Y)$. The upper bound follows directly from the Proposition~\ref{proposition:inequality}. The lower bound can be derived by applying the data processing inequality to the Markov dependency $C \rightarrow X \rightarrow Y \rightarrow C$ that can be confirmed from Eq.~\eqref{eq:Markov_chain}.
    \begin{align}
        &I(C;C) \le I(X;Y) \label{eq18} \\  
        \Rightarrow &H(C)  \le I(X;Y) \label{eq19} 
    \end{align}    
    Note that $I(C;C)$ is the self-information that is equal to $H(C)$. 
\end{proof}

\newcommand{\rulesep}{\vrule height 0.11\textheight}
\begin{figure*}[t!]
    \centering
    \subfloat[$\mathcal{T}_\text{class}^\text{color}$]{
    \includegraphics[height=0.12\textheight]{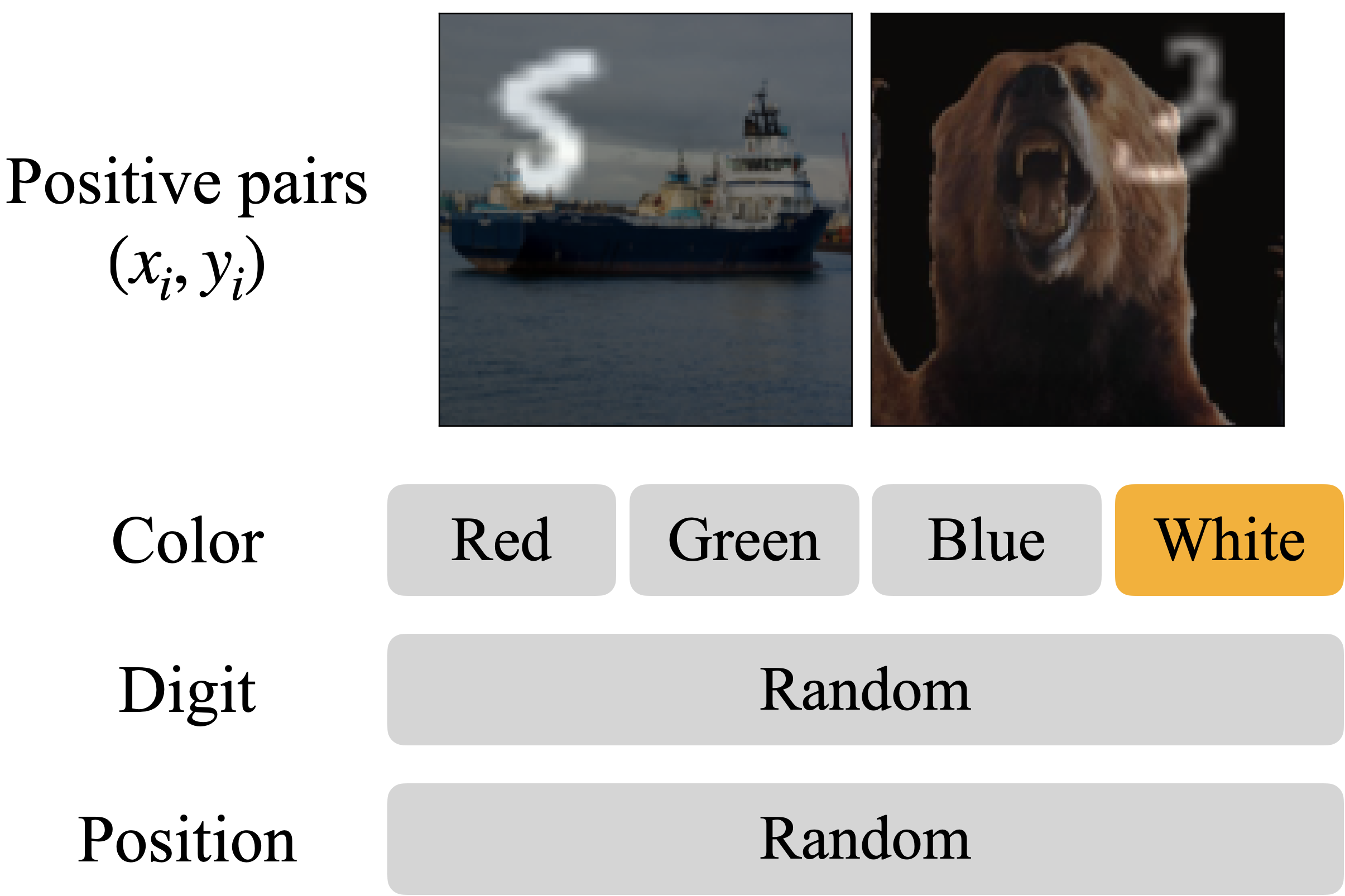}
    }\hspace{0.15cm}\rulesep\hspace{0.1cm}
    \subfloat[$\mathcal{T}_\text{class}^\text{digit}$]{
    \includegraphics[height=0.12\textheight]{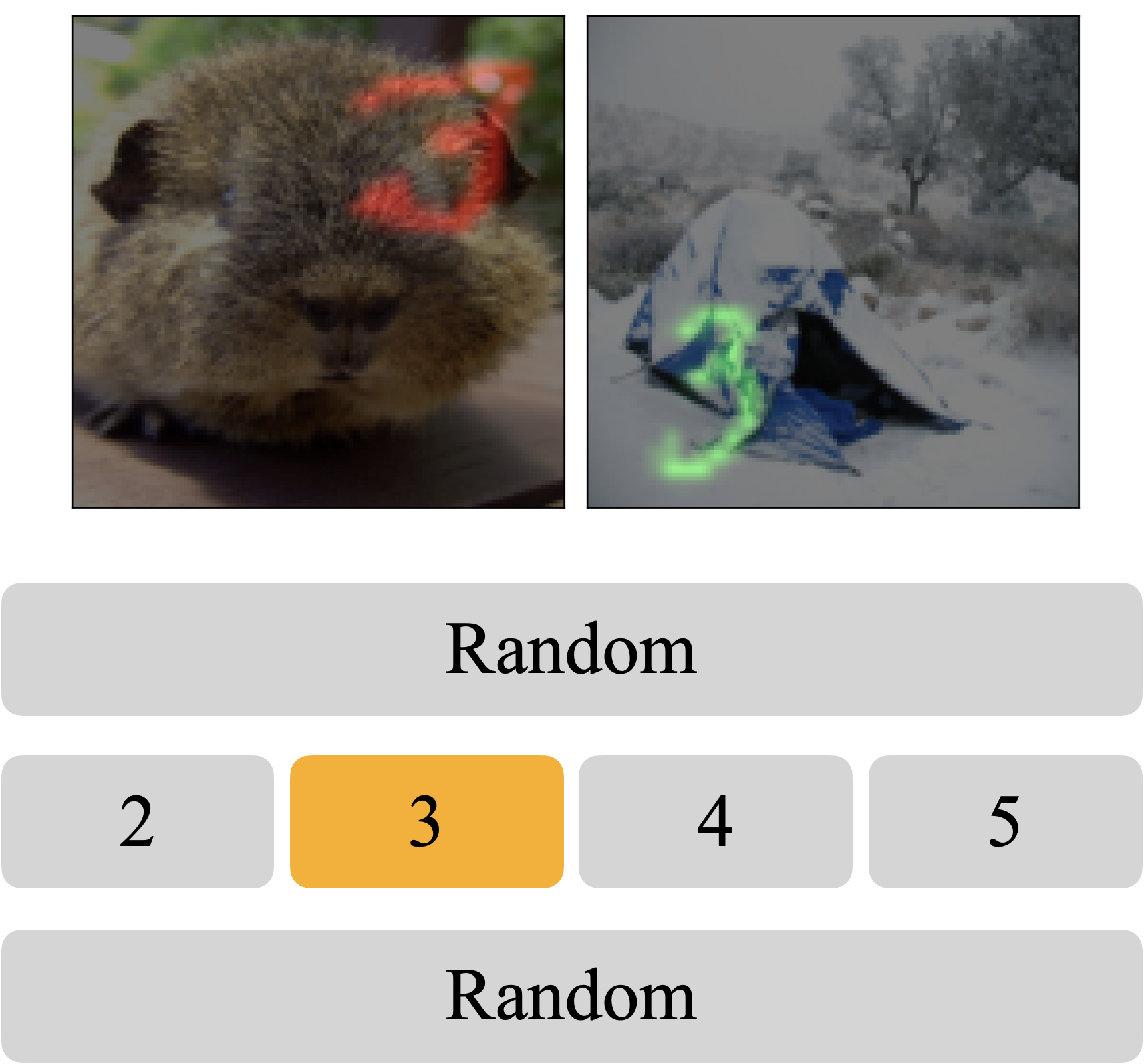}
    }\hspace{0.15cm}\rulesep\hspace{0.1cm}
    \subfloat[$\mathcal{T}_\text{class}^\text{position}$]{
    \includegraphics[height=0.12\textheight]{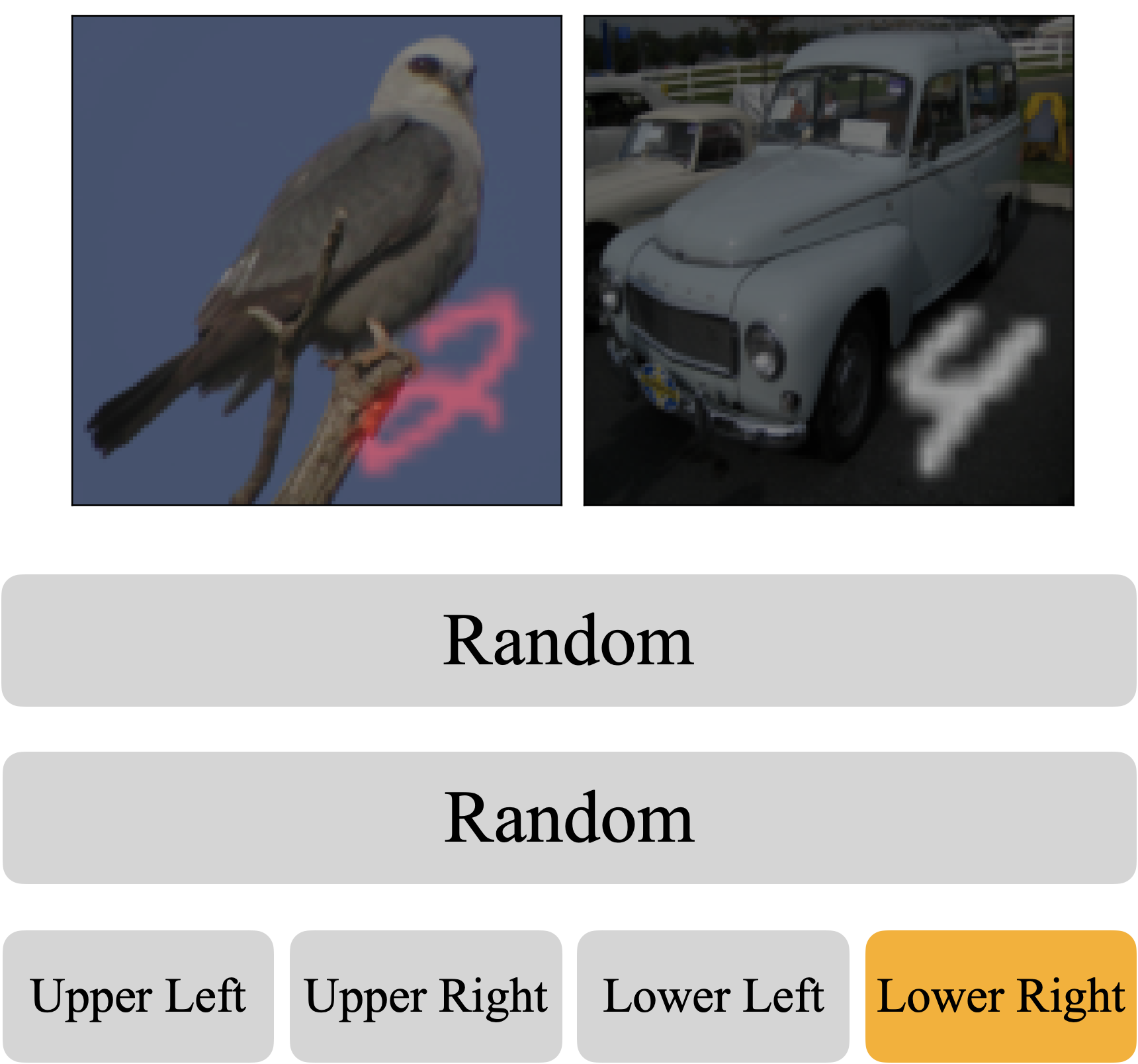}
    }\hspace{0.15cm}\rulesep\hspace{0.1cm}
    \subfloat[$\mathcal{T}_\text{class}^\text{all}$]{
    \includegraphics[height=0.12\textheight]{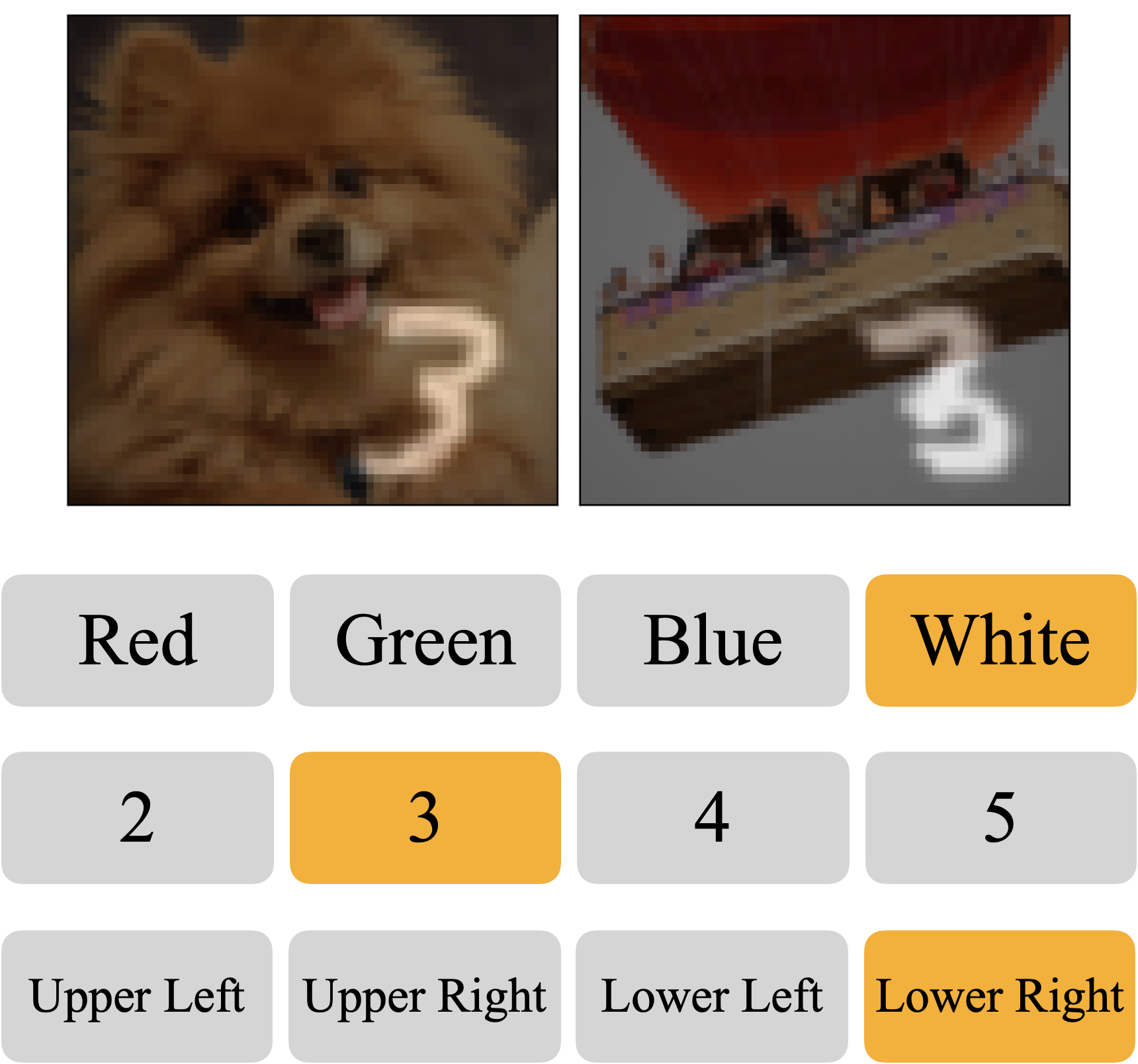}
    }
    \caption{Manipulating class entropy with CDP dataset. When only one of color, digit, and position is consistently matched between the first view and the second view by the same-class sampling~(as shown in (a), (b), and (c)), then the $H(C)$ is $2$ bits~($H(C_\text{color}) = H(C_\text{digit}) = H(C_\text{position}) = 2$ bits). When all three are consistently matched as shown in (d), the $H(C)$ is $6$ bits. By consistently matching only two attributes, $H(C) = 4$ bits can be achieved as well.  
    }
    \label{fig:cdp-dataset}
\end{figure*}

Unlike the same-class sampling, MI of augmentation-based methods such as $\mathcal{T}_\text{SimCLR}$~\cite{chen2020simple}, $\mathcal{T}_\text{AutoAugment}$~\cite{cubuk2018autoaugment}, and $\mathcal{T}_\text{RandAugment}$~\cite{cubuk2020randaugment}
are intractable because the shared information is dependent on the particular choice of $\mathcal{T_\text{aug}}$ whose joint distribution is unknown. 
Among the intractable methods, $\mathcal{T}_\text{SimCLR}$ is studied as the representative example of $\mathcal{T}_\text{aug}$ in our study because it has been most widely adopted in the previous works~\cite{chen2020improved,chen2021exploring,caron2020unsupervised,grill2020bootstrap,zbontar2021barlow,bardes2021vicreg,tomasev2022pushing}.

\subsection{Use of CDP dataset}
\label{subsec:cdp}

The theoretical connection between true MI and $H(C)$ is a convenient property for a rigorous analysis of contrastive learning. To take advantage of the property further, we introduce a synthetic dataset called \textit{CDP dataset} where $H(C)$ can be easily calculated to be $2$, $4$, or $6$ bits. There have been a few similar synthetic datasets in the previous works, but CDP dataset is specifically constructed to facilitate an easy manipulation of $H(C)$. 

In CDP dataset, each source image is constructed by uniformly choosing a color $c_\text{color}$ from $\{ \text{Red}$, $\text{Green}$, $\text{Blue}$, $\text{White} \}$, 
a digit $c_\text{digit}$ from 
$\left\{ 2, 3, 4, 5 \right\}$, 
and a position $c_\text{position}$ from 
$\{$Upper Left, Upper Right, Lower Left, Lower Right$\}$. 
%
The three attributes are independently chosen for each source image. Because of the uniform and independent selection, the entropy of each class label is clearly $H(C_\text{color}) = H(C_\text{digit}) = H(C_\text{position}) = 2$ bits. 

Thanks to the way the CDP dataset is constructed, the $H(C)$ under the same-class sampling can be easily manipulated as shown in Figure~\ref{fig:cdp-dataset}. Note that random ImageNet examples are linearly mixed up in the background to make the dataset realistic. If only the color attribute is consistently chosen between the first view and the second view (i.e., within each positive pair) as shown Figure~\ref{fig:cdp-dataset}(a), it corresponds to a downstream task whose class label is the color information only. In this case, its positive pairing is denoted as $\mathcal{T}_\text{class}^\text{color}$ and the true MI is bounded as $I_\text{class}(X;Y) \le H(C_\text{color})=2$ bits because of Proposition~\ref{proposition:inequality}. Similarly, $I_\text{class}(X;Y) \le 2$ is true for Figure~\ref{fig:cdp-dataset}(b) and Figure~\ref{fig:cdp-dataset}(c). When all three attributes are consistently chosen for each pair as shown in Figure~\ref{fig:cdp-dataset}(d), it corresponds to a downstream task whose class label is the combination of color, digit, and position information. Then, the true MI is bounded as $I_\text{class}(X;Y) \le H(C_\text{all}) = H(C_\text{color}) + H(C_\text{digit}) + H(C_\text{position}) = 6$ bits. Note that the entropies add up because the three attributes are independent. 

Despite the intuitive way of constructing CDP dataset, the strict equality $I_\text{class}(X;Y) = H(C)$ holds only when the condition $\hat{I}_\text{class}(X;Y) = H(C)$ in Theorem~\ref{theorem:equality} is satisfied. However, we found that it is not challenging to encounter such cases as we will show in the MI analysis cases studied in Section~\ref{sec:results}.

We are not the first to propose a synthetic dataset that is analysis-friendly. For instance, similar datasets have been suggested in \cite{tian2020makes,hermann2020shapes}. However, their focus is different from ours where feature suppression or task-dependence of optimal views is studied. The RandBit dataset of \cite{chen2021intriguing} is also similar, but their goal is to study explicit and controllable competing features. The convenient manipulation of $H(C)$ is the design-goal for CDP dataset. We also note that CDP dataset does not have object-centric bias addressed in \cite{purushwalkam2020demystifying} and can be easily controlled by enforcing dependencies among the three information sources.

\subsection{Post-training MI estimation}
\label{subsec:post_training_mi_estimation}

\begin{figure*}[tb!]
    \centering
    \includegraphics[width=\textwidth]{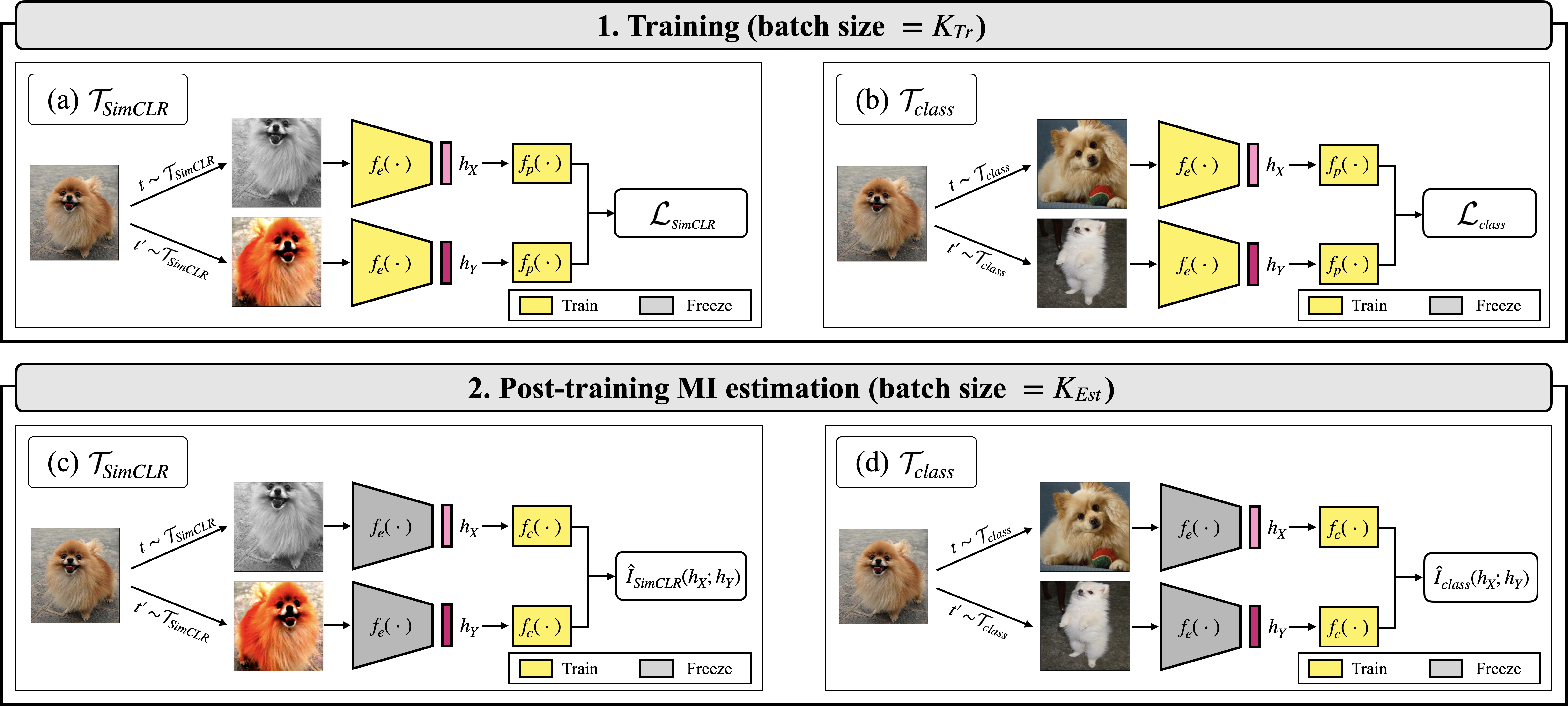}
    \caption{Training vs. post-training MI estimation. 
    (Top) Training: We train the encoder $f_e(\cdot)$ and the projection head $f_p(\cdot)$ to minimize the InfoNCE loss $\mathcal{L}$. 
    (a) With augmentation-based method
    $\mathcal{T}_\text{SimCLR}$. 
    (b) With same-class sampling
    $\mathcal{T}_\text{class}$. 
    (Bottom) Post-training MI estimation: We train the critic $f_c(\cdot)$ to maximize the InfoNCE bound $\hat{I}(h_X;h_Y)$ while $f_e(\cdot)$ is frozen. 
    (c) With augmentation-based method
    $\mathcal{T}_\text{SimCLR}$. 
    (d) With same-class sampling
    $\mathcal{T}_\text{class}$.
    }
    \label{fig:training_and_MIestimation}
\end{figure*}

The value of $\mathcal{L}_\textit{InfoNCE}$ is readily accessible during training because $\mathcal{L}_\textit{InfoNCE}$ is the training loss. This means MI estimation, $\hat{I}(X;Y)$, can be also easily derived by plugging in $\mathcal{L}_\textit{InfoNCE}$ into Eq.~\eqref{eq:definition:estimated_mi} during the training. Because of the straightforward observation, MI estimation has been concurrently performed during training or performed once at the end of training in many existing works. 
This convention, however, can lead to an inaccurate or misleading MI estimation values for two reasons. First, the encoder is being trained with its weights dynamically changing. In this case, it is unclear if a reliable MI estimation can be performed. Second, the hyperparameters are chosen for the purpose of network training and not for the purpose of MI estimation. In particular, the choice of training batch size can be undesirable for MI estimation as we will show in Section~\ref{subsec:4_1_logk_is_a_myth}. 
To address these problems, we propose \textit{post-training MI estimation} where we completely separate the MI estimation phase from the training phase.


Training of contrastive learning is illustrated in the top part of Figure~\ref{fig:training_and_MIestimation}. Because the choice of positive-pairing plays a pivotal role in what is learned, we denote the InfoNCE loss as $\mathcal{L}_\text{SimCLR}$ or $\mathcal{L}_\text{class}$ depending on whether $\augsimclr$ or $\augclass$ is used as the positive pairing. The batch size used for the training is denoted as $K_\text{Tr}$. 

The post-training MI estimation is shown in the bottom part of Figure~\ref{fig:training_and_MIestimation}. Because the goal is an accurate estimation of mutual information, we perform MI estimation after the completion of training. In this way, we can keep the encoder network frozen and only train the critic function after a proper initialization. In our work, we always choose the critic network $f_c(\cdot)$ of post-training MI estimation (shown in the bottom part of Figure~\ref{fig:training_and_MIestimation}) to be the same as the projection header $f_p(\cdot)$ of the contrastive learning (shown in the top part of Figure~\ref{fig:training_and_MIestimation}) where a common MLP network is used. Also, we choose the batch size of MI estimation, $K_\text{Est}$, as a sufficiently large value such that the estimation accuracy does not need to be constrained by the batch size. 

For the post-training MI estimation, we obtain flexibility in critic function and batch size because MI estimation phase is separated from the training phase. In fact, positive-pairing can be flexibly chosen as well. Because we already have a frozen encoder that we want to analyze, we can choose a positive-pairing that is different from the one used in contrastive learning. As we will see in Section~\ref{sec:results}, the choice of positive-pairing for the MI analysis is an important factor because it affects $p(x,y)$, $p(x)p(y)$, and the corresponding mutual information. In our MI analysis in Section~\ref{sec:results}, we investigate $\augsimclr$ and $\augclass$ for post-training MI estimation. When $\augsimclr$ is used, the mutual information of $p_\text{SimCLR}(x,y)$, i.e., $\misimclr$, is being estimated. When $\augclass$ is used, the mutual information of $p_\text{class}(x,y)$, i.e., $\miclass$, is being estimated. 

Post-training MI estimation was developed to improve analysis rigorousness of contrastive learning. However, its applicability does not need to be limited to contrastive learning. It can analyze any given frozen encoder because everything including positive-pairing can be flexibly chosen according to the goal of MI analysis. For instance, any publicly available pre-trained networks can be analyzed including the ones trained in a supervised manner. Therefore, post-training MI estimation can be used for analyzing supervised encoders. We take advantage of this general applicability and compare supervised networks together in Section~\ref{subsec:4_2_MI_and_downstream_performance}.

Finally, our focus is in the MI analysis of the encoded representations and not in the raw input images. Therefore, we are interested in the encoded representations $h_X$ and $h_Y$ and not in the input images $X$ and $Y$. While all the derivations so far have used the notations of $X$ and $Y$, they are also applicable to $h_X$ and $h_Y$ because they are either general results for two random variables or because their extension to $h_X$ and $h_Y$ is straightforward thanks to the Markov dependencies $X\rightarrow h_X$ and $Y\rightarrow h_Y$.

\section{Three Cases of MI Analysis}
\label{sec:results}

In this section, we present MI analysis results of three cases in contrastive learning. The cases are related to fundamental problems that have been partly addressed in the existing literature, and we mainly focus on simplifying the existing knowledge and developing a deeper understanding. The three methods in Section~\ref{sec:method} are heavily utilized, demonstrating how to improve rigorousness in MI analysis.

\begin{figure*}[tb!]
    \centering
    \captionsetup[subfigure]{labelformat=empty}
    \subfloat[]{
        \resizebox{0.48\textwidth}{!}{%
        \begin{tabular}{|p{1.35cm}>{\centering\arraybackslash}p{0.75cm}>{\centering\arraybackslash}p{0.75cm}>{\centering\arraybackslash}p{0.75cm}>{\centering\arraybackslash}p{0.75cm}>{\centering\arraybackslash}p{0.75cm}>{\centering\arraybackslash}p{0.75cm}>{\centering\arraybackslash}p{0.75cm}>{\centering\arraybackslash}p{0.75cm}|}\hline
            $K_\text{Tr}$ & 2 & 4 & 8 & 16 & 32 & 64 & 128 & 256 \\ \hline
            Acc. (\%) & 98.8 & \textbf{99.8} & 99.7 & 99.7 & 99.6 & 99.4 & 99.4 & 99.2 \\ \hline
        \end{tabular}
        }
    }\hspace{-0.15cm}
    \subfloat[]{
        \resizebox{0.48\textwidth}{!}{%
        \begin{tabular}{|p{1.35cm}>{\centering\arraybackslash}p{0.75cm}>{\centering\arraybackslash}p{0.75cm}>{\centering\arraybackslash}p{0.75cm}>{\centering\arraybackslash}p{0.75cm}>{\centering\arraybackslash}p{0.75cm}>{\centering\arraybackslash}p{0.75cm}>{\centering\arraybackslash}p{0.75cm}>{\centering\arraybackslash}p{0.75cm}|}\hline
            $K_\text{Tr}$ & 2 & 4 & 8 & 16 & 32 & 64 & 128 & 256 \\ \hline
            Acc. (\%) & 96.0 & \textbf{99.7} & \textbf{99.7} & \textbf{99.7} & 99.5 & 99.2 & 99.0 & 98.6 \\ \hline
        \end{tabular}
        }
    } \\ \vspace{-0.77cm}
    \subfloat[(a) ResNet-18]{
    \hspace{0.308cm}
    \includegraphics[width=0.45\textwidth]{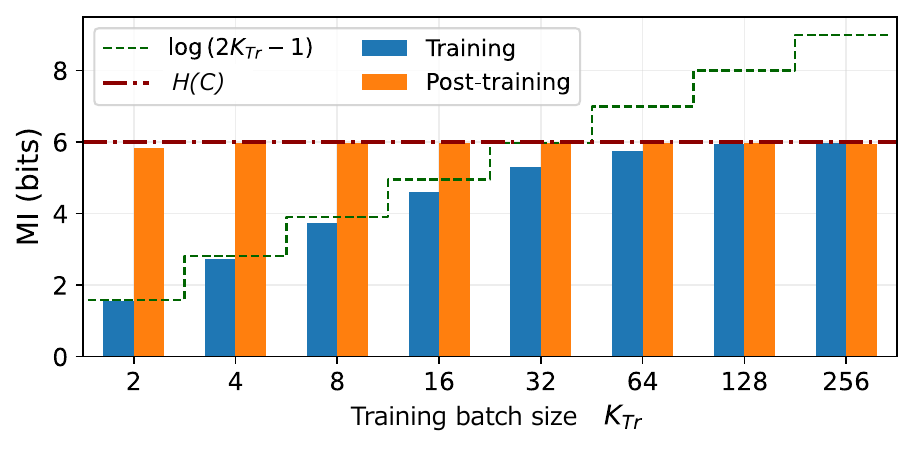}}
    \subfloat[(b) ResNet-50]{
    \hspace{0.55cm}
    \includegraphics[width=0.45\textwidth]{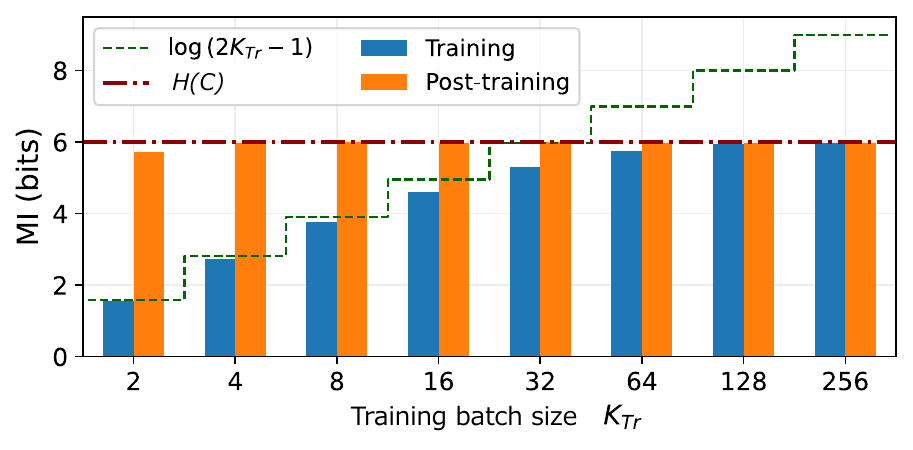}}
    \vspace{-0.1cm}
    \caption{
    $H(C)$, training MI, and post-training MI: For CDP dataset with the downstream task of classifying all of color, digit, and position ($H(C)=6$ bits), we perform contrastive learning of two ResNet models using $\mathcal{L}_\text{class}$ as the loss (as in Figure~\ref{fig:training_and_MIestimation}(b)). We evaluate MI estimation of training~(blue) and post-training MI estimation~(orange). For post-training MI estimation, $K_\text{Est}=256$ was used. 
    }
    \label{fig:log_k}
\end{figure*}

\subsection{Small batch size is not harmful}
\label{subsec:4_1_logk_is_a_myth}

\begin{redbox}{Subject for investigation \#1:}
    A small batch size is undesirable for contrastive learning because of InfoNCE's $\mathcal{O}(\log{K})$ bound~\cite{hjelm2018infomax,tian2020contrastive,bachman2019learning,chen2020simple,sordoni2021decomposed,wu2020conditional,song2020mlcpc}.
\end{redbox}
\vspace{-0.1cm}
\begin{greenbox}{MI analysis result \#1:}
    A small batch size limits the range of training loss, but in general it neither restrains the information in the learned representation nor bounds the downstream-task performance.
\end{greenbox}

A straightforward interpretation of contrastive learning is to understand it as a process of maximizing mutual information between the first view $h_X$ and the second view $h_Y$. Meanwhile, the $\log{(2K-1)}$ bound in Proposition~\ref{proposition:infonce_2Kminus1_inequality} has been well known, where usually $K=K_\text{Tr}$ is implicitly assumed~\cite{oord2018cpc,sordoni2021decomposed,mcallester2020formal,poole2019variational}. Because of the straightforward interpretation and the $\log{(2K_\text{Tr}-1)}$ bound, it has been frequently speculated that a small batch size limits the mutual information and therefore negatively affects the contrastive learning. To overcome the speculated limitation, many of the previous works have either increased the batch size~\cite{hjelm2018infomax,tian2020contrastive,bachman2019learning} or modified the InfoNCE loss~\cite{sordoni2021decomposed,wu2020conditional,song2020mlcpc}. 

Using our methods in Section~\ref{sec:method}, we can design an experiment and clearly point out the incorrect parts of the speculation. Using the CDP dataset and $\miclass$, we have performed two sets of experiments as shown in Figure~\ref{fig:log_k}. In the experiments, we have evaluated $H(C)$, the in-training MI estimation, and the post-training MI estimation. The in-training MI estimation (i.e., estimated MI of training) was evaluated by re-using the training setup including the training batch size $K_\text{Tr}$ and by converting the training loss into to the in-training MI using Eq.~\eqref{eq:definition:estimated_mi}. Therefore, it is always upper bounded by $\log{(2K_\text{Tr}-1)}$ of Proposition~\ref{proposition:infonce_2Kminus1_inequality}. In contrast, the post-training MI estimation is not upper bounded by $\log{(2K_\text{Tr}-1)}$ because a larger batch size of $K_\text{Est}=256$ was used for the estimation. In fact, the resulting $\miclass$ is almost constant at 6 bits regardless of the choice of $K_\text{Tr}$ that was used for the encoder's contrastive learning. In this case, $\miclass\approx 6 \text{bits}=H(C)$ and therefore we can conclude that the true MI value is also around 6 bits by Theorem~\ref{theorem:equality} and that the post-training MI estimation is almost exact. 

Based on the experiments, we can draw two conclusions on the speculation. First, the limiting effect on the in-training MI does not necessarily limit the true MI of the encoder. A small $K_\text{Tr}$ merely means that the learning dynamics will be affected by the choice and there is no reason for the encoder's inherent encoding capability to be limited by $\log{(2K_\text{Tr}-1)}$. In fact, the downstream task's linear evaluation performance is always very high regardless of the choice of $K_\text{Tr}$ -- between 96.0\% and 99.8\% as shown in the top parts of Figure~\ref{fig:log_k}.  
Second, there is no reason to prefer either a small $K_\text{Tr}$ or a large $K_\text{Tr}$ from the viewpoint of mutual information. In the experiments, the best performance occurs for $K_\text{Tr}=4$ for ResNet-18 and $K_\text{Tr}=4,8,16$ for ResNet-50. 

The achievability of high performance with a small $K_\text{Tr}$ is not new. 
For instance,~\cite{yeh2021decoupled} proposed decoupled contrastive learning and showed its effectiveness with a small $K_\text{Tr}$ on CIFAR and ImageNet datasets. Also, a small $K_\text{Tr}$ for learning sentence embedding was studied in~\cite{gao2021simcse,yan2021consert,chen2022information}. 
Compared to the previous studies, our analysis clearly points out that MI estimation of in-training should not be confused with the true MI of the encoder. The existing confusion related to the $\log{(2K_\text{Tr}-1)}$ speculation is due to the failure to differentiate the estimated MI of training and the true MI of the encoder. Our investigation is based on all thee methods described in Section~\ref{sec:method} where Theorem~\ref{theorem:equality} plays a crucial role for understanding MI in contrastive learning.

\subsection{MI stands as a superior measure of representation quality}
\label{subsec:4_2_MI_and_downstream_performance}

\begin{table*}[tb!]
\centering
    \caption{Post-training MI estimation results for ResNet-50(Pretrained) on ImageNet-100 and ImageNet-1k. %
    Sixteen pre-trained models in \citet{goyal2021vissl,rw2019timm,khosla2020supervised} are used to evaluate the effectiveness of $\misimclr$ and $\miclass$.
    }
    \vspace{-0.1cm}
    \label{tab:4.1.resnet50}
    \resizebox{0.95\textwidth}{!}{%
    \begin{tabular}{lcccccc}
    \toprule
    \multirow{2}{*}{Algorithm} & \multicolumn{3}{c}{\textit{ImageNet-100}} & \multicolumn{3}{c}{\textit{ImageNet-1k}} \\ \cmidrule(lr){2-4}\cmidrule(lr){5-7}
    & Acc. (\%) & $\misimclr$ & $\miclass$ & Acc. (\%) & $\misimclr$ & $\miclass$ \\\midrule
    SupCon~\cite{khosla2020supervised} & 94.40 & 7.889 & 6.100 & 78.72 & 8.722 & 7.783 \\
    Supervised pretrained & 93.00 & 7.598 & 5.816 & 74.11 & 8.378 & 6.761 \\
    SwAV~\cite{caron2020unsupervised} & 92.52 & 8.541 & 5.560 & 74.78 & 9.428 & 6.214 \\
    DeepCluster-v2~\cite{caron2020unsupervised} & 92.38 & 8.540 & 5.559 & 73.65 & 9.416 & 6.232 \\
    DINO~\cite{caron2021dino} & 92.22 & 8.443 & 5.539 & 74.22 & 9.313 & 6.133 \\
    Barlow Twins~\cite{zbontar2021barlow} & 90.80 & 8.528 & 5.513 & 72.82 & 9.407 & 6.157 \\
    PIRL~\cite{misra2020self} & 90.58 & 8.584 & 5.480 & 70.51 & 9.481 & 6.247 \\
    SeLa-v2~\cite{caron2020unsupervised} & 89.50 & 6.020 & 5.039 & 69.66 & 7.354 & 5.774 \\
    SimCLR~\cite{chen2020simple} & 89.40 & 8.669 & 5.546 & 69.12 & 9.580 & 6.277 \\
    MoCo-v2~\cite{chen2020improved} & 87.54 & 8.592 & 5.490 & 63.89 & 9.499 & 6.221 \\
    NPID++~\cite{misra2020self} & 79.60 & 8.190 & 4.792 & 56.60 & 9.009 & 4.692 \\
    MoCo~\cite{he2020momentum} & 76.94 & 8.338 & 4.904 & 47.05 & 9.155 & 4.907 \\
    NPID~\cite{wu2018unsupervised} & 76.68 & 8.039 & 4.188 & 52.70 & 8.821 & 3.836 \\
    ClusterFit~\cite{yan2020clusterfit} & 75.66 & 8.016 & 4.155 & 48.81 & 8.773 & 3.915 \\
    RotNet~\cite{gidaris2018unsupervised} & 66.90 & 7.020 & 2.916 & 41.54 & 7.696 & 2.802 \\
    Jigsaw~\cite{noroozi2016unsupervised} & 56.74 & 6.339 & 2.510 & 30.85 & 7.155 & 2.583 \\
    \midrule
    Pearson's correlation coefficient $\rho$ with Acc. & & 0.510 & 0.967 & & 0.535 & 0.943 \\
    Kendall's rank correlation coefficient $\tau_K$ with Acc. & & 0.233 & 0.883 & & 0.233 & 0.617 \\ \bottomrule
    \end{tabular}
    }
\end{table*}

\begin{table*}[tb!]
    \centering
    \caption{Summary of Pearson's correlation and Kendall's rank correlation for six practical scenarios including the scenario in Table~\ref{tab:4.1.resnet50}. The experiment of Table~\ref{tab:4.1.resnet50} was repeated over five additional scenarios where encoders and datasets were varied. Only the resulting correlation values are presented for the brevity. In addition to $\misimclr$ and $\miclass$, alignment, uniformity, and tolerance are investigated together. For all metrics, $+1.000$ corresponds to the largest correlation with the downstream performance. For alignment and uniformity, $-1$ was multiplied because a smaller value corresponds to a better representation quality.
    }
    \vspace{-0.1cm}
    \label{tab:4.1.all}
        \begin{tabular}{lllccccc}
            \toprule
            \multirow{2}{*}{\shortstack[l]{Correlation \\ Coefficient}} & \multirow{2}{*}{Encoder} & \multirow{2}{*}{Dataset} & \multicolumn{5}{c}{Metrics} \\\cmidrule(lr){4-8}
             & & & Alignment & Uniformity & Tolerance & $\misimclr$ & $\miclass$ \\
            \midrule\midrule
            \multirow{8}{*}{\shortstack[l]{Pearson's \\ correlation \\ coefficient $\rho$ \\ with \\ linear accuracy}} &
            $\phantom{A}$ResNet-$\left\{18,50\right\}$ & CIFAR-10 & $\phantom{-}$0.035 & $-$0.218 & $\phantom{-}$0.046 & $-$0.041 & \textbf{0.634} \\
             & $\phantom{A}$ResNet-$\left\{18,50\right\}$ & ImageNet-100 & $\phantom{-}$0.250 & $-$0.173 & $\phantom{-}$0.247 & $\phantom{-}$0.085 & \textbf{0.805} \\
             & $\phantom{A}$ResNet-50(Pretrained)         & ImageNet-100 & $-$0.197 & $\phantom{-}$0.381 & $-$0.200 & $\phantom{-}$0.510 & \textbf{0.967} \\
             & $\phantom{A}$ResNet-50(Pretrained)         & ImageNet-1k & $\phantom{-}$0.012 & $\phantom{-}$0.213 & $\phantom{-}$0.007 & $\phantom{-}$0.535 & \textbf{0.943} \\
             & $\phantom{A}$ViT(Pretrained)               & ImageNet-100 & $-$0.442 & $\phantom{-}$0.623 & $-$0.444 & $\phantom{-}$0.727 & \textbf{0.975} \\
             & $\phantom{A}$ViT(Pretrained)            & ImageNet-1k & $-$0.360 & $\phantom{-}$0.546 & $-$0.360 & $\phantom{-}$0.790 & \textbf{0.979} \\ \cmidrule(lr){2-8}
             & \multicolumn{2}{c}{Average} & $\phantom{-}$0.036 & $\phantom{-}$0.188 & $\phantom{-}$0.036 & $\phantom{-}$0.231 & \textbf{0.899} \\ 
             & \multicolumn{2}{c}{Worst} & $-$0.442 & $-$0.218 & $-$0.444 & $-$0.041 & \textbf{0.634} \\ \midrule \midrule
            \multirow{8}{*}{\shortstack[l]{Kendall's \\ rank correlation \\ coefficient $\tau_K$ \\ with \\ linear accuracy}} & $\phantom{A}$ResNet-$\left\{ 18, 50 \right\}$ & CIFAR-10 & $\phantom{-}$0.067 & $-$0.067 & $\phantom{-}$0.138 & $-$0.067 & \textbf{0.467} \\
             & $\phantom{A}$ResNet-$\left\{ 18, 50 \right\}$ & ImageNet-100 & $-$0.067 & $\phantom{-}$0.333 & $-$0.067 & $\phantom{-}$0.067 & \textbf{0.467} \\
             & $\phantom{A}$ResNet-50(Pretrained) & ImageNet-100 & $\phantom{-}$0.159 & $\phantom{-}$0.092 & $\phantom{-}$0.159 & $\phantom{-}$0.233 & \textbf{0.883} \\
             & $\phantom{A}$ResNet-50(Pretrained) & ImageNet-1k & $\phantom{-}$0.250 & $-$0.033 & $\phantom{-}$0.250 & $\phantom{-}$0.233 & \textbf{0.617} \\
             & $\phantom{A}$ViT(Pretrained) & ImageNet-100 & $-$0.026 & $\phantom{-}$0.308 & $-$0.026 & $\phantom{-}$0.513 & \textbf{0.821} \\
             & $\phantom{A}$ViT(Pretrained) & ImageNet-1k & $-$0.055 & $\phantom{-}$0.418 & $-$0.055 & $\phantom{-}$0.600 & \textbf{0.891} \\ \cmidrule(lr){2-8}
             & \multicolumn{2}{c}{Average} & $\phantom{-}$0.116 & $\phantom{-}$0.159 & $\phantom{-}$0.126 & $\phantom{-}$0.122 & \textbf{0.670} \\
             & \multicolumn{2}{c}{Worst} & $-$0.067 & $-$0.067 & $-$0.067 & $-$0.067 & \textbf{0.467} \\\bottomrule
        \end{tabular}
\end{table*}

\begin{redbox}{Subject for investigation \#2:}
    Large MI is not predictive of downstream performance~\cite[\S3.1]{tschannen2019mutual}.
\end{redbox}
\vspace{-0.1cm}
\begin{greenbox}{MI analysis result \#2:}
    MI can be a superior measure for practical networks when an appropriate positive pairing $\mathcal{T}$ is chosen and a post-training MI estimation is applied. 
\end{greenbox}


The main goal of contrastive learning is to learn desirable representations without making use of side information such as annotations. Evaluation of the representation quality, however, has been a challenging problem. In the absence of a clear method for analytical evaluation, downstream performance of a set of benchmarks has been used as a proxy where the linear evaluation in \cite{alain2016understanding} is utilized to limit the capacity of the evaluation network. As the second subject, we investigate if mutual information can be a meaningful measure of representation quality. 



As laid out in Eq.~\eqref{eq:definition:estimated_mi} of Definition~\ref{definition:estimated_mi}, the InfoNCE loss is directly connected to the InfoNCE estimation of MI. Because of Eq.~\eqref{eq:definition:estimated_mi}, it had been commonly believed that a larger MI should result in a better downstream performance because it directly means a smaller InfoNCE loss. The belief, however, had been neither theoretically proven nor empirically grounded and eventually it was empirically shown to be incorrect. In~\cite{tschannen2019mutual}, the 
authors have designed an experiment where an invertible RealNVP architecture~\cite{dinh2016density} is used as the encoder and the top-bottom split MNIST is used as the two views $X$ and $Y$. For the experiment, the true MI of the representation remains constant throughout the training because of the encoder's invertibility -- an invertible function preserves the true MI~\cite{cover1999elements}. However, it was empirically shown that the in-training MI estimation actually varies as the training proceeds, which clearly shows the estimated MI's inaccuracy. The authors also observes that the downstream accuracy is correlated with the estimated MI and points out that the correlation is clearly misleading because the estimated MI is not a reliable reflection of the true MI -- what really matters is the true MI that remains constant in the experiment. An additional experiment with an adversarial training was provided to show that an inverse correlation can be demonstrated, too.  Overall,~\cite{tschannen2019mutual} clearly showed that the downstream performance can be manipulated while the true MI remains constant.


%
%

While the result of~\cite{tschannen2019mutual} was enlightening, the experiment in~\cite{tschannen2019mutual} can be also misleading for establishing practical insights -- the invertible networks and top-bottom split MNIST dataset are hardly used in practice because they do not yield a high-performance encoder. 
In particular, it is unclear if the joint distribution $p_{\text{aug}}(x,y)$ of the top-bottom split MNIST is sufficiently related to the downstream tasks of interest. Therefore, we focus on an MI analysis with two realistic considerations. The first is the network. We focus on high-performance or historically well-known networks. Networks trained with both supervised and unsupervised methods are included. The second is the positive pairing scheme. We consider $\mathcal{T}_\text{SimCLR}$ that is broadly used and $\mathcal{T}_\text{class}$ that is clearly related to the downstream tasks.


The post-training MI estimation introduced in Section~\ref{subsec:post_training_mi_estimation} enables a wide applicability because any pre-trained network, regardless of its training method, can be considered as a frozen encoder for the purpose of MI analysis. Taking advantage of the post-training MI estimation, we have  devised an experiment to examine MI as a measure of representation quality. In the experiment, we have investigated a total of $16$ pre-trained networks and have evaluated Pearson's and Kendall's rank correlations between the downstream performance and MI. We have included a variety of networks trained with supervised, contrastive, non-contrastive, and pretext learning. 
The results are shown in Table~\ref{tab:4.1.resnet50} where the downstream performance is shown together with $\misimclr$ and $\miclass$. For $\misimclr$, the Kendall's rank correlation with the downstream performance is low ($0.233$ and $0.233$) but the Pearson's correlation is reasonably high ($0.510$ and $0.535$). Considering that $16$ networks including several high-performance networks are analyzed, the Pearson's correlation can be interpreted as a supportive evidence of $\misimclr$ being a reasonable measure of representation quality. For $\miclass$, all the correlation values are very large where the Pearson's correlation turns out to be almost $1.000$ ($0.967$ and $0.943$). Considering that a variety of practical networks are included in the $16$, the result indicates that $\miclass$ is an outstanding measure of representation quality. 

In contrast to $\misimclr$ that is an unsupervised measure, $\miclass$ is a supervised measure where it is dependent on the same-class sampling. Therefore, it is crucial to understand if the superiority of $\miclass$ is mostly owing to its supervised nature and not MI itself. To inspect if non-MI measures can outperform $\miclass$ if same-class sampling is used as in $\miclass$, we have investigated alignment~\cite{wang2020understanding}, uniformity~\cite{wang2020understanding}, and tolerance~\cite{wang2021understanding}. They are very popular measures of representation quality. The experiment in Table~\ref{tab:4.1.resnet50} was expanded to include five additional scenarios, and the correlation summary for the six scenarios is provided in Table~\ref{tab:4.1.all} (the full details can be found in Appendix~\ref{supp:full_results_4.1}). The result is rather surprising where the three popular measures fail to correlate well, even though their evaluation also utilized same-class sampling where class label information is fully utilized. We have also investigated their unsupervised versions using SimCLR augmentation, $\mathcal{T}_\text{SimCLR}$, and obtained similar random and low correlations as in the same-class sampling results shown in Table~\ref{tab:4.1.all}. 

The outstanding performance of $\miclass$ as a measure of representation quality leads us to a few interesting conclusions. First, MI can be a practical and superior measure because it can correlate well with the downstream performance of practical and high-performance networks. Second, the choice of $\mathcal{T}$ for MI estimation can be critical. When $\mathcal{T}$ is chosen carefully such that its associated $p(x,y)$ is sufficiently related to the downstream task of interest, MI can be a reliable measure of representation quality. Note that the top-bottom split MNIST in~\cite{tschannen2019mutual} might not be a desirable choice because its associated $p(x,y)$ and the learned encoder might not be sufficiently related to the commonly considered benchmarks and their downstream tasks. Third, the popular measures such as alignment, uniformity, and tolerance should be interpreted with a caution. Our experiment results reveal that it can be misleading to assume that a stronger value is indicative of a better representation quality. Considering that a trade-off between alignment and uniformity exists~\cite{wang2020understanding}, perhaps it is natural that each individual measure fails. A more complex measure that combines multiple individual measures might work better, but such a study is beyond the scope of our study. 
Finally, our empirical study can be expanded to the recent theoretical developments in contrastive learning where analytical bounds have been derived to connect contrastive learning with supervised learning~\cite{arora2019theoretical,nozawa2021understanding,ash2021investigating,bao2022surrogate}. When we examined the analytical bounds as representation measures, however, even the analytically tightest bound resulted in Pearson's correlation of $-0.409$ and Kendall's rank correlation of $-0.182$.
%
%
Most recently, limitations of the theoretical results including their dependency on properties of augmentations were addressed in~\cite{saunshi2022understanding}.




\subsection{Discrimination between task-relevant information and task-irrelevant information is difficult}
\label{subsec:infomin}

\begin{redbox}{Subject for investigation \#3:}
    In contrastive learning, task-irrelevant information needs to be discarded for a better generalization~\cite{tian2020makes, wang2022rethinking, tsai2020self, xiao2020should, chen2021intriguing}.
\end{redbox}
\vspace{-0.1cm}
\begin{greenbox}{MI analysis result \#3:}
    Discriminating task-irrelevant information from task-relevant information can be challenging. Furthermore, task-irrelevant information source does not necessarily harm the generalization of the downstream task.
\end{greenbox}

The choice of augmentation is known to determine the type of invariance that is learned during contrastive learning~\cite{tian2020makes,tsai2020self,xiao2020should,chen2021intriguing}. 
In particular, \citet{tian2020makes} formalized this idea into the InfoMin principle: 
`good views for a given downstream task in contrastive learning should retain task-relevant information while minimizing irrelevant nuisances'.
Many of the subsequent works considered the InfoMin as a fundamental principle for data augmentation design where discarding task-irrelevant information is commonly believed to be helpful. In the InfoMin and related works, MI estimation has been used to analyze task-relevant information, task-irrelevant information, and minimal sufficient representation.

\begin{figure}[t!]
    \centering
    \subfloat[Color-jittering]{
    \includegraphics[width=0.8\linewidth]{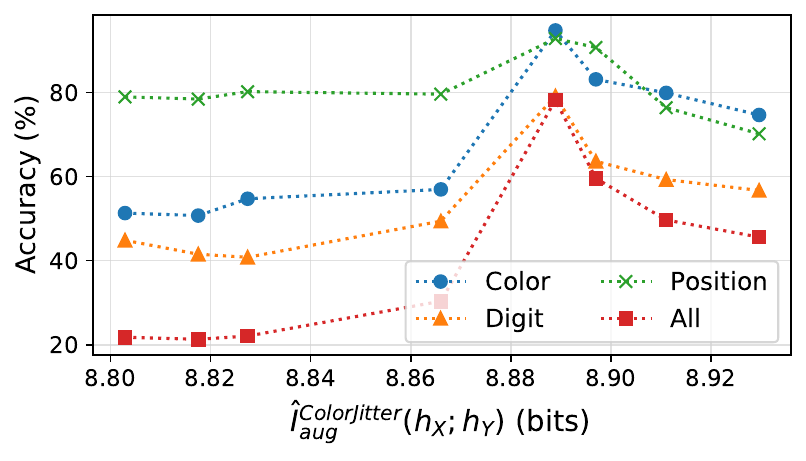}}\hfill
    \subfloat[Random-resized-crop]{
    \includegraphics[width=0.8\linewidth]{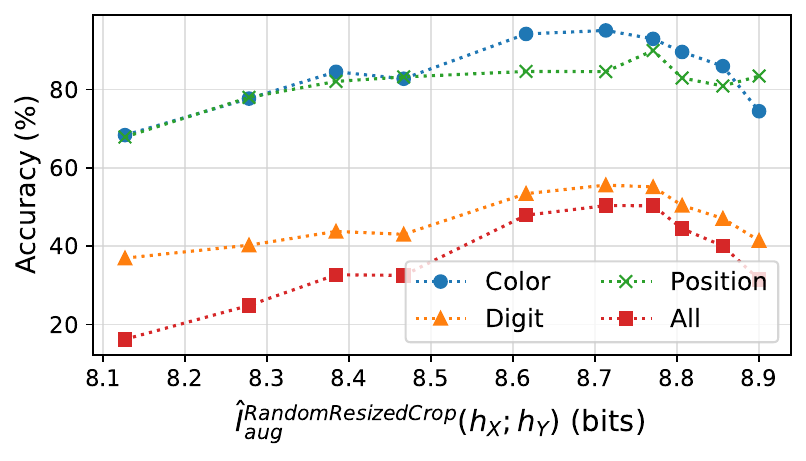}}
    \caption{
    Linear evaluation performance when (a) only color-jittering or (b) only random-resized-crop is used as the augmentation of contrastive learning. The same augmentation as in the training was used for the post-training MI estimation. Experiments were conducted with CDP dataset and ResNet-50. 
    }
    \label{fig:general.aug}
\end{figure}

We re-examine an existing MI analysis to develop a deeper understanding. 
The key concept in~\cite{tian2020makes} is the reverse-U shape of performance curve when plotted against mutual information. The basic idea is that too small MI indicates task-relevant information is lacking and too large MI indicates task-irrelevant information is excessive. Therefore, the performance peaks when MI is appropriately adjusted and thus resulting in a reverse-U shape for the performance curve. As a main MI analysis, the control of MI is achieved by adjusting the strength of color-jittering augmentation or the strength of random-resized-crop augmentation. We have repeated the analysis using CDP dataset and the results are shown in Figure~\ref{fig:general.aug}. We have generated four performance curves for four downstream tasks of $C_\text{color}$, $C_\text{digit}$, $C_\text{position}$, and $C_\text{all}$. If we look at only one single curve, say digit shown in orange, it follows the original result of reverse-U shape~\cite[Figure~5(a)]{tian2020makes} and thus possibly supporting the InfoMin principle. If we look into the four curves more carefully, however, we can identify issues that undermine the original analysis. 

In Figure~\ref{fig:general.aug}(a), all four curves manifest reverse-U shape and they all peak at the same MI value. This indicates that the performance of all four downstream tasks is simultaneously maximized when the strength of color-jittering augmentation is adjusted to a particular level. 
If we consider the three individual tasks of $C_\text{color}$, $C_\text{digit}$, and $C_\text{position}$, their class labels were independently generated with three independent information sources. Therefore, we can say that an individual task's task-relevant information is the other individual tasks' task-irrelevant information. Given that the task-relevant information and task-irrelevant information are different for the three tasks, the simultaneous peaking is very unlikely to be owing to the InfoMin. 
In fact, even a stronger point can be made for the color task $C_\text{color}$.
The color-jittering augmentation is only relevant to color and it is not directly related to position or digit. Therefore, we would expect the curve of color to be different from the curves of position or digit, but all three peak simultaneously. 
Furthermore, color-jittering-based contrastive learning is supposed to reduce the color information in the learned representation. Because of the straightforward relationship, a monotonically increasing or decreasing curve is expected instead of the reverse-U shape.  
Overall, the justification of InfoMin principle based on the observation of reserve-U shape does not appear to be sound. Similar conclusions can be drawn for the random-resized-crop augmentation shown in Figure~\ref{fig:general.aug}(b).

\begin{figure} [t!]
    \centering
    \subfloat[ResNet-18]{
    \includegraphics[width=0.48\linewidth]{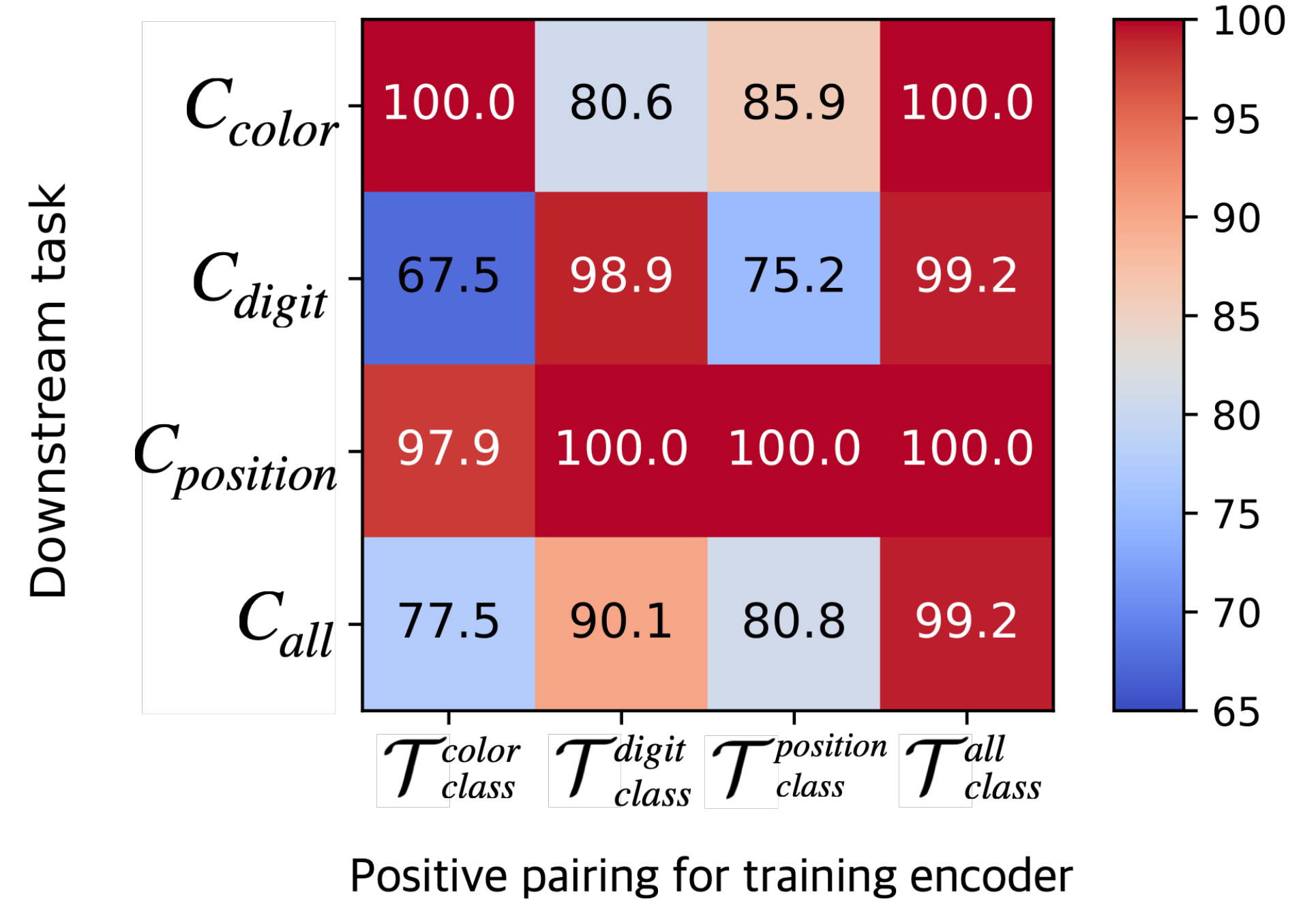}}\hfill
    \subfloat[ResNet-50]{
    \includegraphics[width=0.48\linewidth]{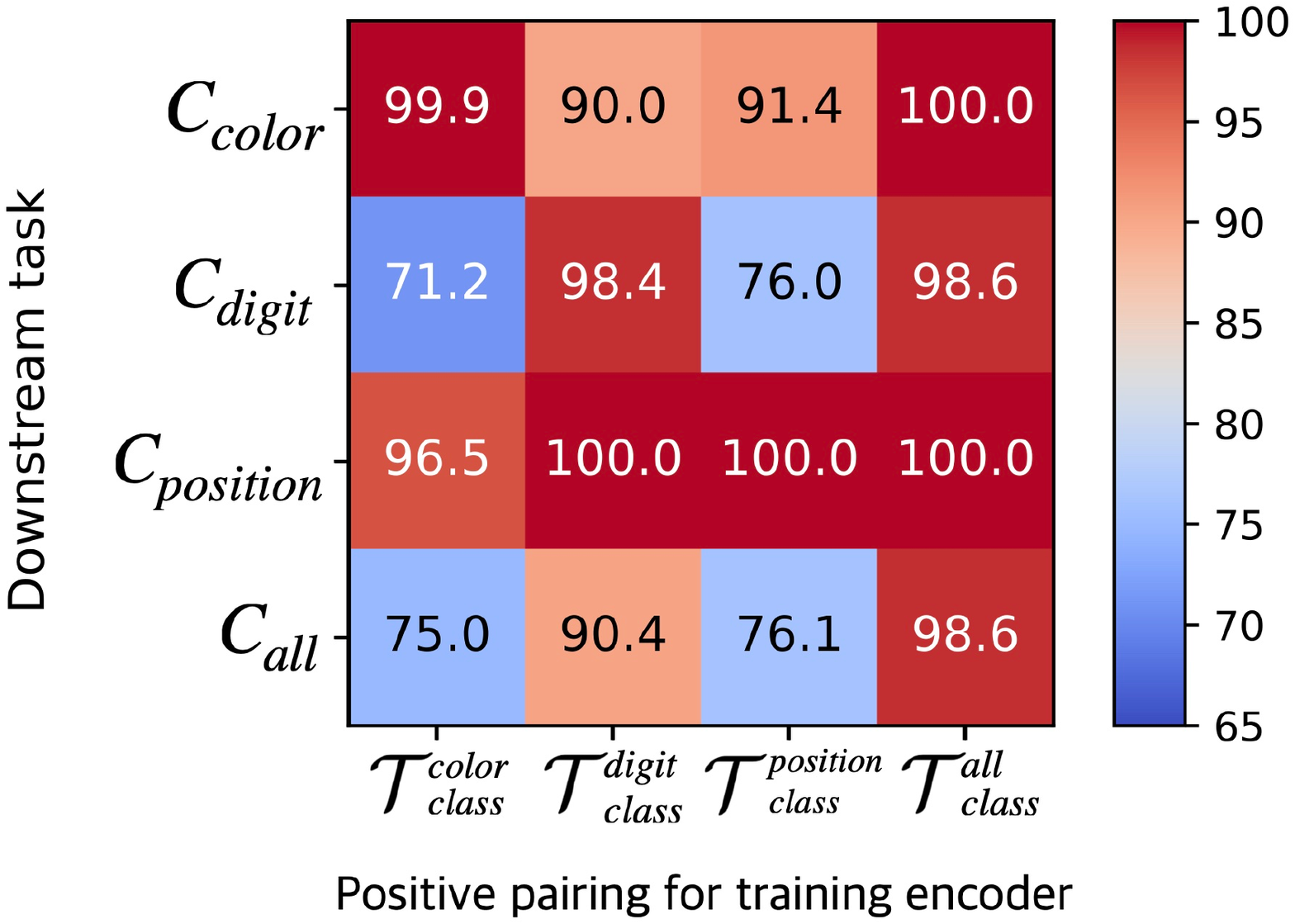}}
    \caption{Linear evaluation performance of CDP dataset for task-dependent training. The task in $x$-axis indicates the positive pairing $\mathcal{T}$ used for training. The task in $y$-axis indicates the evaluated downstream task $C$. 
    }
    \label{fig:aug.sup}
\end{figure}

To understand the observed reverse-U shape better, we carried out a comprehensive experiment using the CDP dataset. We have experimented with all of the four strategies for positive pairing of training (i.e.,  $\mathcal{T}_\text{class}^\text{color}$, $\mathcal{T}_\text{class}^\text{digit}$, $\mathcal{T}_\text{class}^\text{position}$, and $\mathcal{T}_\text{class}^\text{all}$) and also for downstream task of evaluation (i.e., $C_\text{color}$, $C_\text{digit}$, $C_\text{position}$, and $C_\text{all}$). 
The results are shown in Figure~\ref{fig:aug.sup} for ResNet-18 and ResNet-50. 
It can be noticed that the diagonal elements achieve a high performance as foreseen by InfoMin principle. 
The non-diagonal elements, however, exhibit a few incompatible behaviors.
First, the performance of all four tasks is high when $\mathcal{T}_\text{class}^\text{all}$ is used for training. In particular, the three individual tasks $C_\text{color}$, $C_\text{digit}$, and $C_\text{position}$ do not suffer despite the apparent presence of the task-irrelevant information. This observation is directly against the InfoMin principle.
Second, the performance for downstream task $C_\text{position}$ is always high regardless of the augmentation chosen for training. This might be because position information is always needed for decoding color or digit information - even though the information sources are independent, the position information might have become task-relevant because of the way raw image is constructed or how the information is processed by the given encoder. If true, this means that even independent information sources can become interwoven in a complex way. 
The corresponding post-training MI estimation shows a similar pattern, and the results are provided Figure~\ref{fig:aug.mi} in Appendix B.


%


\section{Discussion}
\label{sec:discussions}

We provide two points of discussion in this section. 

\subsection{MI as a learning objective} 
\label{sec:discussion_measure_vs_learingObjective}

MI can be an excellent measure of representation quality as shown in Section~\ref{subsec:4_2_MI_and_downstream_performance}. But then, can it be also an excellent choice as a learning objective? To better understand the role of MI as a learning objective, we have compared four learning objectives. Two of them are unsupervised where 
$\mathcal{L}_\text{SimCLR}$ represents the standard implementation of contrastive learning and $\mathcal{L}_\text{aug,best-known}$ represents the best known customization for each task. The other two are supervised where $\mathcal{L}_\text{class}$ is the basic implementation that utilizes the downstream task information only and   $\mathcal{L}_\text{SupCon}$~\cite{khosla2020supervised} is a customized implementation that combines $\mathcal{L}_\text{class}$ with SimCLR-like augmentations using an empirical selection. 

The comparison results are summarized in Table~\ref{tab:loss}. For the commonly used benchmarks of CIFAR-10, ImageNet-100, and ImageNet-1k, two aspects can be observed to conclude that even MI of the downstream task, corresponding to $\mathcal{L}_\text{class}$, is not sufficient when it is used alone. 
First, we can compare the supervised $\mathcal{L}_\text{class}$ with the two unsupervised learning objectives. It can be observed that $\mathcal{L}_\text{class}$ performs better than the standard $\mathcal{L}_\text{SimCLR}$ for all three cases. However, it performs worse than $\mathcal{L}_\text{aug,best-known}$ for two out of three cases. Considering that $\mathcal{L}_\text{aug,best-known}$ is unsupervised, the result indicates that \textit{a full supervision with downstream-task information only is not sufficient} and even a carefully customized unsupervised learning can outperform the basic supervised representation learning.
Second, when the basic supervised loss $\mathcal{L}_\text{class}$ is strengthened with customized augmentation techniques as in $\mathcal{L}_\text{SupCon}$~\citep{khosla2020supervised}, it can outperform $\mathcal{L}_\text{class}$ as shown in Table~\ref{tab:loss}. This shows that \textit{supervised learning can be improved with additional learning signals even in the presence of the full downstream-task information}~\citep{khosla2020supervised}. As expected, supervised $\mathcal{L}_\text{SupCon}$ also outperforms unsupervised $\mathcal{L}_\text{aug,best-known}$.


The two observations clearly show that the downstream task information alone is not sufficient for the best learning. In machine learning, the advantage of a properly chosen surrogate loss has been well understood. As in the example of cross-entropy loss used for learning binary classification, often a learning objective that differs from the ultimate performance objective results in a better learning. The same applies to contrastive learning. While it is a necessary condition for a high-performance network to have a large amount of downstream-task information contained in the representation as we have seen in Section~\ref{subsec:4_2_MI_and_downstream_performance}, the \textit{training} of such a network can benefit from a carefully chosen surrogate loss. 

Recently, the limitation of MI-based contrastive learning has become clear. Many of the recent works have developed non-contrastive learning methods that can outperform MI-based contrastive learning~\citep{caron2020unsupervised,grill2020bootstrap,zbontar2021barlow,bardes2021vicreg}. Also, it was shown that a small modification in the loss function of contrastive learning can be beneficial~\citep{yeh2021decoupled}, indicating that the loss function's deviation from the exact MI formulation can be advantageous depending on the task of interest. In~\cite{chen2021intriguing}, it was shown that differences between contrastive losses are small with a deep projection head.
Finally, we show in Appendix C 
that the viewpoint of \textit{Noise Contrastive Estimation}~(NCE)~\citep{gutmann2010noise,oord2018cpc} with a prudent choice for the noise distribution (i.e., negative sample distribution in contrastive learning) might be more adequate for explaining the success of unsupervised representation learning than the viewpoint of MI maximization.

\begin{table}[tb!]
    \centering
    \caption{Comparison of linear evaluation performance for a set of loss functions. 
    }
    \label{tab:loss}
    \vspace{0.1cm}
    \resizebox{0.45\textwidth}{!}{%
    \begin{tabular}{lllll}
    \toprule
    \multirow{2}{*}{Loss} & Unsupervised & Unsupervised & Supervised & Supervised \\
     & $\mathcal{L}_\text{SimCLR}$ & $\mathcal{L}_\text{aug,best-known}$ & $\mathcal{L}_\text{class}$ & $\mathcal{L}_\text{SupCon}$  \\\midrule
    CIFAR-10 & 93.0 & 94.1 (SWD~\citep{chen2021intriguing}) & 93.1 & \textbf{96.0}~\cite{khosla2020supervised} 
    \\ 
    ImageNet-100 & 77.8 & 84.5 (MoCo-v2+MoCHi~\citep{kalantidis2020hard}) & 87.4 & \textbf{87.6} \\ 
    ImagNet-1k & 69.1 (SimCLR~\citep{chen2020simple}) & 76.4 (HCA~\citep{xu2020seed}) & 75.2  & \textbf{78.7}~\cite{khosla2020supervised} \\ \bottomrule  
    %
    %
    \end{tabular}
    }
\end{table}

\subsection{Towards a further advancement in unsupervised representation learning}
\label{sec:discussion_truly_unsupervised_learning}


In contrastive learning, an annotated validation dataset is utilized for multiple purposes including encoder selection, learning method refinement, and augmentation design. Given the vital role of validation dataset in contrastive learning, it can be puzzling to notice that contrastive learning is categorized as an unsupervised method -- the validation dataset inherently embodies downstream-task information. Apparently, the validation dataset does not directly affect the network parameters because it is not used in the process of gradient descent. Instead, it indirectly affects the network parameters, especially through the augmentation design. The choice of augmentation design is dependent on the validation performance. The chosen augmentation defines the joint distribution $p_\text{aug}(x,y)$. In turn, $p_\text{aug}(x,y)$ decides the MI of learning that directly affects the network parameters. From this viewpoint, contrastive learning is not purely unsupervised. Then, why do we consider contrastive learning to be an unsupervised method? 

Prior to the success of contrastive learning, a variety of pretext learning techniques had been developed~\cite{doersch2015unsupervised, pathak2016context, noroozi2016unsupervised, gidaris2018unsupervised}. They also relied on a validation dataset for designing their pretext techniques, but contrastive learning turned out to be far more superior in terms of \textit{performance}. Contrastive learning is also much \textit{easier to develop}. For any given dataset and task, usually a set of effective augmentation techniques are already known for their supervised learning. Then, such known supervised augmentation techniques can be easily refined to form an outstanding $\mathcal{T_\text{aug}}$ for contrastive learning. Furthermore, the learned representation of contrastive learning tends to \textit{generalize better} over diverse downstream tasks when compared to the existing pretext methods.
{The transition from existing pretext learning to contrastive learning represents a significant milestone in the field of unsupervised representation learning. As we move forward, further advancement beyond contrastive learning remains as an essential goal.
We note that a variety of efforts are on-going within the fields of visual and multi-modal representation learning. For instance, masking was adopted for visual unsupervised learning in~\cite{he2022masked} where the use of validation dataset was limited to the tuning of masking ratio.}

\vspace{1.5cm}

\IEEEraisesectionheading{
\section{Conclusion}
\label{sec:Conclusion}}

Mutual information has emerged as a pivotal factor in shaping the landscape of unsupervised contrastive learning. However, when employed as an analytical tool, the challenge of accurately estimating mutual information can result in interpretations that lack the desired level of rigor.
In this study, we present a collection of straightforward yet impactful methods aimed at enhancing the robustness of mutual information analysis. Through three illustrative case studies, we showcase the practical utility of these methods.
Additionally, we discuss two fundamental issues. The first pertains to the inherent limitations of mutual information when employed as a learning objective. The second is regarding the rationale behind the pursuit of an unsupervised learning approach that surpasses contrastive learning, particularly in terms of the downstream applicability of the acquired representations. 
Through our study, we provide insights into the broader context of contrastive learning and mutual information analysis and lay the groundwork for potential avenues of progress within the field of unsupervised representation learning.



\bibliographystyle{unsrtnat}
\bibliography{reference}

\clearpage

\onecolumn
\appendix

\section*{A. Full results for Table~\ref{tab:4.1.all}}
\label{supp:full_results_4.1}

We provide the full results used for generating the summary shown in Table~\ref{tab:4.1.all}. When we had to train a network from the scratch, we tried three different temperature parameters for each dataset.

\begin{table*}[!ht]
\caption{Metrics evaluated on CIFAR-10 dataset using ResNet-18 and ResNet-50 models.}
\centering
\resizebox{\textwidth}{!}{%
\begin{tabular}{lcccccccc}
    \toprule
    Model & Temperature & Acc. (\%) & Alignment $\downarrow$ & Uniformity $\downarrow$ & Tolerance $\uparrow$ & Contrastive loss $\downarrow$ & $\hat{I}_\text{SimCLR}(X;Y)$ & $\hat{I}_\text{class}(X;Y)$ \\\midrule
    ResNet-18 & 0.1 & 90.03 & 1.170 & -2.622 & 0.415 & 0.642 & 8.063 & 2.717 \\
    ResNet-18 & 0.3 & 91.11 & 1.345 & -3.074 & 0.327 & 3.489 & 7.912 & 2.874 \\
    ResNet-18 & 0.5 & 90.97 & 1.129 & -2.758 & 0.434 & 4.509 & 7.730 & 2.756 \\
    ResNet-50 & 0.1 & 92.06 & 1.034 & -2.223 & 0.483 & 0.476 & 8.117 & 2.806 \\
    ResNet-50 & 0.3 & 92.97 & 1.346 & -2.922 & 0.327 & 3.323 & 7.954 & 2.902 \\
    ResNet-50 & 0.5 & 93.01 & 1.071 & -2.402 & 0.467 & 4.489 & 7.879 & 2.803 \\ \midrule
    \multicolumn{3}{l}{
    Pearson's $\rho$ with Acc.}
     & 0.035 & -0.218 & 0.046 & -0.367 & -0.041 & \textbf{0.634} \\
    \multicolumn{3}{l}{
    Kendall's $\tau_K$ with Acc.}
     & 0.067 & -0.067 & 0.138 & -0.067 & -0.067 & \textbf{0.467} \\\bottomrule
\end{tabular}
}
\end{table*}

\begin{table*}[!ht]
\caption{Metrics evaluated on ImageNet-100 dataset using ResNet-18 and ResNet-50 models.}
\centering
\resizebox{\textwidth}{!}{%
\begin{tabular}{lcccccccc}
    \toprule
    Model & Temperature & Acc. (\%) & Alignment $\downarrow$ & Uniformity $\downarrow$ & Tolerance $\uparrow$ & Contrastive loss $\downarrow$ & $\hat{I}_\text{SimCLR}(X;Y)$ & $\hat{I}_\text{class}(X;Y)$ \\\midrule
    ResNet-18 & 0.1 & 72.60 & 0.919 & -2.475 & 0.546 & 0.209 & 8.367 & 3.394 \\
    ResNet-18 & 0.2 & 76.42 & 0.950 & -2.872 & 0.529 & 1.529 & 8.378 & 3.907 \\
    ResNet-18 & 0.3 & 75.66 & 0.823 & -2.597 & 0.591 & 2.680 & 8.316 & 3.857 \\ 
    ResNet-50 & 0.1 & 74.08 & 0.127 & -0.328 & 0.938 & 0.179 & 8.414 & 3.967 \\
    ResNet-50 & 0.2 & 75.52 & 0.117 & -0.354 & 0.942 & 3.738 & 8.350 & 4.186 \\
    ResNet-50 & 0.3 & 77.80 & 0.180 & -0.536 & 0.911 & 2.650 & 8.406 & 4.263 \\\midrule
    \multicolumn{3}{l}{
    Pearson's $\rho$ with Acc.}
     & 0.250 & -0.173 & 0.247 & -0.672 & 0.085 & \textbf{0.805} \\
    \multicolumn{3}{l}{
    Kendall's $\tau_K$ with Acc.}
     & -0.067 & 0.333 & -0.067 & -0.200 & 0.067 & \textbf{0.467} \\\bottomrule
\end{tabular}
}
\end{table*}

We additionally test a variety of pre-trained models loaded from \cite{goyal2021vissl,khosla2020supervised,rw2019timm}. We inspect 16 pre-trained ResNet-50 models and 13 pre-trained ViT models. All models are pre-trained by ImageNet-1k dataset. We load the pre-trained models and evaluate the linear accuracy and the metrics. The results are shown below.

\begin{table*}[!ht]
\caption{Metrics evaluated on ImageNet-100 dataset using pre-trained ResNet-50 models.}
\centering
\resizebox{0.8\textwidth}{!}{%
\begin{tabular}{lcccccc}
    \toprule
    Algorithm       & Acc. (\%) & Alignment $\downarrow$ & Uniformity $\downarrow$ & Tolerance $\uparrow$ & $\hat{I}_\text{SimCLR}(X;Y)$ & $\hat{I}_\text{class}(X;Y)$ \\\midrule
    SupCon~\cite{khosla2020supervised} & 94.40 & 0.791 & -2.943 & 0.610 & 7.889 & 6.100 \\
    Supervised pretrained & 93.00 & 1.189 & -3.196 & 0.415 & 7.598 & 5.816 \\
    SwAV~\cite{caron2020unsupervised} & 92.52 & 0.725 & -1.937 & 0.643 & 8.544 & 5.560 \\
    DeepCluster-v2~\cite{caron2020unsupervised} & 92.38 & 0.603 & -1.585 & 0.703 & 8.540 & 5.559 \\
    DINO~\cite{caron2021dino} & 92.22 & 0.822 & -2.135 & 0.596 & 8.443 & 5.539 \\
    Barlow Twins~\cite{zbontar2021barlow} & 90.80 & 1.040 & -2.685 & 0.487 & 8.528 & 5.513 \\
    PIRL~\cite{misra2020self} & 90.58 & 1.107 & -3.445 & 0.458 & 8.584 & 5.480 \\
    SeLa-v2~\cite{caron2020unsupervised} & 89.50 & 0.578 & -1.547 & 0.717 & 6.020 & 5.039 \\
    SimCLR~\cite{chen2020simple} & 89.40 & 1.132 & -3.320 & 0.445 & 8.669 & 5.546 \\
    MoCo-v2~\cite{chen2020improved} & 87.54 & 0.845 & -2.940 & 0.584 & 8.592 & 5.490 \\
    NPID++~\cite{misra2020self} & 79.60 & 1.283 & -3.170 & 0.371 & 8.190 & 4.792 \\
    MoCo~\cite{he2020momentum} & 76.94 & 1.189 & -3.196 & 0.415 & 8.338 & 4.904 \\
    NPID~\cite{wu2018unsupervised} & 76.68 & 1.122 & -2.650 & 0.449 & 8.039 & 4.188 \\
    ClusterFit~\cite{yan2020clusterfit} & 75.66 & 1.292 & -3.113 & 0.366 & 8.016 & 4.155 \\
    RotNet~\cite{gidaris2018unsupervised} & 66.90 & 0.816 & -2.019 & 0.599 & 7.020 & 2.916 \\
    Jigsaw~\cite{noroozi2016unsupervised} & 56.74 & 0.227 & -0.541 & 0.889 & 6.339 & 2.510 \\
    \midrule
    \multicolumn{2}{l}{
    Pearson's $\rho$ with Acc.}
    & -0.197 & 0.381 & -0.200 & 0.510 & \textbf{0.967} \\
    \multicolumn{2}{l}{
    Kendall's $\tau_K$ with Acc.}
    & 0.159 & 0.092 & 0.159 & 0.233 & \textbf{0.883} \\ \bottomrule
    \end{tabular}
}
\end{table*}

\begin{table*}[!ht]
\caption{Metrics evaluated on ImageNet-1k dataset using pre-trained ResNet-50 models.}
\centering
\resizebox{0.8\textwidth}{!}{%
\begin{tabular}{lcccccc}
    \toprule
    Algorithm & Acc. (\%) & Alignment $\downarrow$ & Uniformity $\downarrow$ & Tolerance $\uparrow$ & $\hat{I}_\text{SimCLR}(X;Y)$ & $\hat{I}_\text{class}(X;Y)$ \\\midrule
    SupCon~\cite{khosla2020supervised} & 78.72 & 0.801 & -2.961 & 0.605 & 8.722 & 7.783 \\
    Supervised pretrained & 74.11 & 0.534 & -1.918 & 0.735 & 8.378 & 6.761 \\
    SwAV~\cite{caron2020unsupervised} & 74.78 & 0.728 & -1.944 & 0.641 & 9.428 & 6.214 \\
    DeepCluster-v2~\cite{caron2020unsupervised} & 73.65 & 0.604 & -1.590 & 0.701 & 9.416 & 6.232 \\
    DINO~\cite{caron2021dino} & 74.22 & 0.822 & -2.136 & 0.594 & 9.313 & 6.133 \\
    Barlow Twins~\cite{zbontar2021barlow} & 72.82 & 1.034 & -2.692 & 0.490 & 9.407 & 6.157 \\
    PIRL~\cite{misra2020self} & 70.51 & 1.119 & -3.476 & 0.451 & 9.481 & 6.247 \\
    SeLa-v2~\cite{caron2020unsupervised} & 69.66 & 0.578 & -1.571 & 0.715 & 7.354 & 5.774 \\
    SimCLR~\cite{chen2020simple} & 69.12 & 1.114 & -3.332 & 0.450 & 9.580 & 6.277 \\
    MoCo-v2~\cite{chen2020improved} & 63.89 & 0.854 & -2.963 & 0.580  & 9.499 & 6.221 \\
    NPID++~\cite{misra2020self} & 56.60 & 1.282 & -3.180 & 0.371 & 9.009 & 4.692 \\
    MoCo~\cite{he2020momentum} & 47.05 & 1.196 & -3.210 & 0.414 & 9.155 & 4.907 \\
    NPID~\cite{wu2018unsupervised} & 52.70 & 1.116 & -2.640 & 0.453 & 8.821 & 3.836 \\
    ClusterFit~\cite{yan2020clusterfit} & 48.81 & 1.286 & -3.101 & 0.369 & 8.773 & 3.915 \\
    RotNet~\cite{gidaris2018unsupervised} & 41.54 & 0.807 & -1.973 & 0.603 & 7.696 & 2.802 \\
    Jigsaw~\cite{noroozi2016unsupervised} & 30.85 & 0.221 & -0.532 & 0.891 & 7.155 & 2.583 \\
    \midrule
    \multicolumn{2}{l}{
    Pearson's $\rho$ with Acc.}
    & 0.012 & 0.213 & 0.007 & 0.535 & \textbf{0.943} \\
    \multicolumn{2}{l}{
    Kendall's $\tau_K$ with Acc.} & 0.250 & -0.033 & 0.250 & 0.233 & \textbf{0.617} \\ \bottomrule
    \end{tabular}
}
\end{table*}

\begin{table*}[!ht]
    \centering
    \caption{Metrics evaluated on ImageNet-100 dataset using pre-trained ViT models.}
    \resizebox{0.8\textwidth}{!}{%
        \begin{tabular}{lcccccc}
        \toprule
        Algorithm 
        & Acc. (\%) & Alignment $\downarrow$ & Uniformity $\downarrow$ & Tolerance $\uparrow$ & $\hat{I}_\text{SimCLR}(X;Y)$ & $\hat{I}_\text{class}(X;Y)$ \\\midrule
        Swin-B~\cite{liu2021swin} & 96.20 & 1.022 & -3.660 & 0.498 & 8.073 & 6.222 \\
        Supervised pretrained (ViT-B/16)~\cite{dosovitskiy2020vit} & 95.36 & 1.122 & -3.834 & 0.450 & 8.252 & 5.977 \\
        PiT-B~\cite{heo2021pit} & 94.62 & 1.028 & -3.686 & 0.496 & 7.895 & 6.398 \\
        DeiT (ViT-B/16)~\cite{touvron2021deit} & 94.30 & 1.006 & -3.754 & 0.507 & 7.799 & 6.287 \\
        CaiT (XXS-36/16)~\cite{touvron2021cait} & 93.90 & 0.891 & -3.738 & 0.564 & 7.492 & 5.795 \\
        PiT-S~\cite{heo2021pit} & 93.76 & 1.038 & -3.756 & 0.491 & 7.664 & 6.151 \\
        CaiT (XXS-24/16)~\cite{touvron2021cait} & 93.28 & 0.963 & -3.777 & 0.529 & 7.488 & 5.690 \\
        MoCo(v3) (ViT-B/16)~\cite{chen2021empirical}
        & 93.12 & 0.413 & -1.274 & 0.797 & 8.594 & 5.654 \\
        DINO (ViT-B/16)~\cite{caron2021dino}
        & 92.84 & 1.072 & -3.604 & 0.475 & 8.454 & 5.675 \\
        Supervised pretrained (ViT-S/16)~\cite{dosovitskiy2020vit} & 92.70 & 1.007 & -3.476 & 0.506 & 6.863  & 5.515 \\
        DeiT (ViT-T/16)~\cite{touvron2021deit} & 90.12 & 1.081 & -3.806 & 0.469 & 7.186 & 5.365 \\
        Supervised pretrained (ViT-T/16)~\cite{dosovitskiy2020vit} & 80.14 & 1.154 & -3.208 & 0.436 & 4.988 & 3.814 \\
        DINO (ViT-S/16)~\cite{caron2021dino} & 76.54 & 0.256 & -0.662 & 0.874 & 6.868 & 3.525 \\
        \midrule
        \multicolumn{2}{l}{
        Pearson's $\rho$ with Acc.}
         & -0.442 & 0.623 & -0.444 & 0.727 & \textbf{0.975} \\
        \multicolumn{2}{l}{
        Kendall's $\tau_K$ with Acc.}
         & -0.026 & 0.308 & -0.026 & 0.513 & \textbf{0.821} \\ \bottomrule
        \end{tabular}
        }
\end{table*}

\begin{table*}[!ht]
    \centering
        \caption{Metrics evaluated on ImageNet-1k dataset using pre-trained ViT models. Because of the computational burden, the two largest models are excluded.}
    \resizebox{0.8\textwidth}{!}{%
        \begin{tabular}{lcccccc}
        \toprule
        Algorithm 
        & Acc. (\%) & Alignment $\downarrow$ & Uniformity $\downarrow$ & Tolerance $\uparrow$ & $\hat{I}_\text{SimCLR}(X;Y)$ & $\hat{I}_\text{class}(X;Y)$ \\\midrule
        Supervised pretrained (ViT-B/16)~\cite{dosovitskiy2020vit} & 78.93 & 1.159 & -3.888 & 0.432 & 9.199 & 7.208 \\
        DeiT (ViT-B/16)~\cite{touvron2021deit} & 78.34 & 1.054 & -3.833 & 0.483 & 8.679 & 8.009 \\
        PiT-S~\cite{heo2021pit} & 76.81 & 1.078 & -3.832 & 0.471 & 8.513 & 7.543 \\
        CaiT (XXS-36/16)~\cite{touvron2021cait} & 75.67 & 0.918 & -3.839 & 0.550 & 8.373 & 6.795 \\
        MoCo(v3) (ViT-B/16)~\cite{chen2021empirical} & 75.51 & 0.424 & -1.297 & 0.792 & 9.524 & 6.658 \\
        CaiT (XXS-24/16)~\cite{touvron2021cait} & 74.09 & 0.989 & -3.863 & 0.515 & 8.315 & 6.547 \\
        DINO (ViT-B/16)~\cite{caron2021dino} & 73.28 & 1.092 & -3.646 & 0.464 & 9.367 & 6.598 \\
        Supervised pretrained (ViT-S/16)~\cite{dosovitskiy2020vit} & 72.85 & 1.034 & -3.505 & 0.493 & 7.572 & 6.233 \\
        DeiT (ViT-T/16)~\cite{touvron2021deit} & 68.67 & 1.098 & -3.871 & 0.462 & 7.874 & 5.883 \\
        Supervised pretrained (ViT-T/16)~\cite{dosovitskiy2020vit} & 53.01 & 1.147 & -3.199 & 0.437 & 5.474 & 3.741 \\
        DINO (ViT-S/16)~\cite{caron2021dino} & 51.11 & 0.304 & -0.758 & 0.851 & 7.426 & 3.316 \\
        \midrule
        \multicolumn{2}{l}{
        Pearson's $\rho$ with Acc.}
         & -0.360 & 0.546 & -0.360 & 0.790 & \textbf{0.979} \\
        \multicolumn{2}{l}{
        Kendall's $\tau_K$ with Acc.}
         & -0.055 & 0.418 & -0.055 & 0.600 & \textbf{0.891} \\\bottomrule
        \end{tabular}
        }
\end{table*}

\clearpage 

\section*{B. Post-training MI estimation results for Figure~\ref{fig:aug.sup}}
\label{supp:aug.mi}

\begin{figure}[hb!]
    \centering
    \subfloat[ResNet-18]{
    \includegraphics[width=0.35\linewidth]{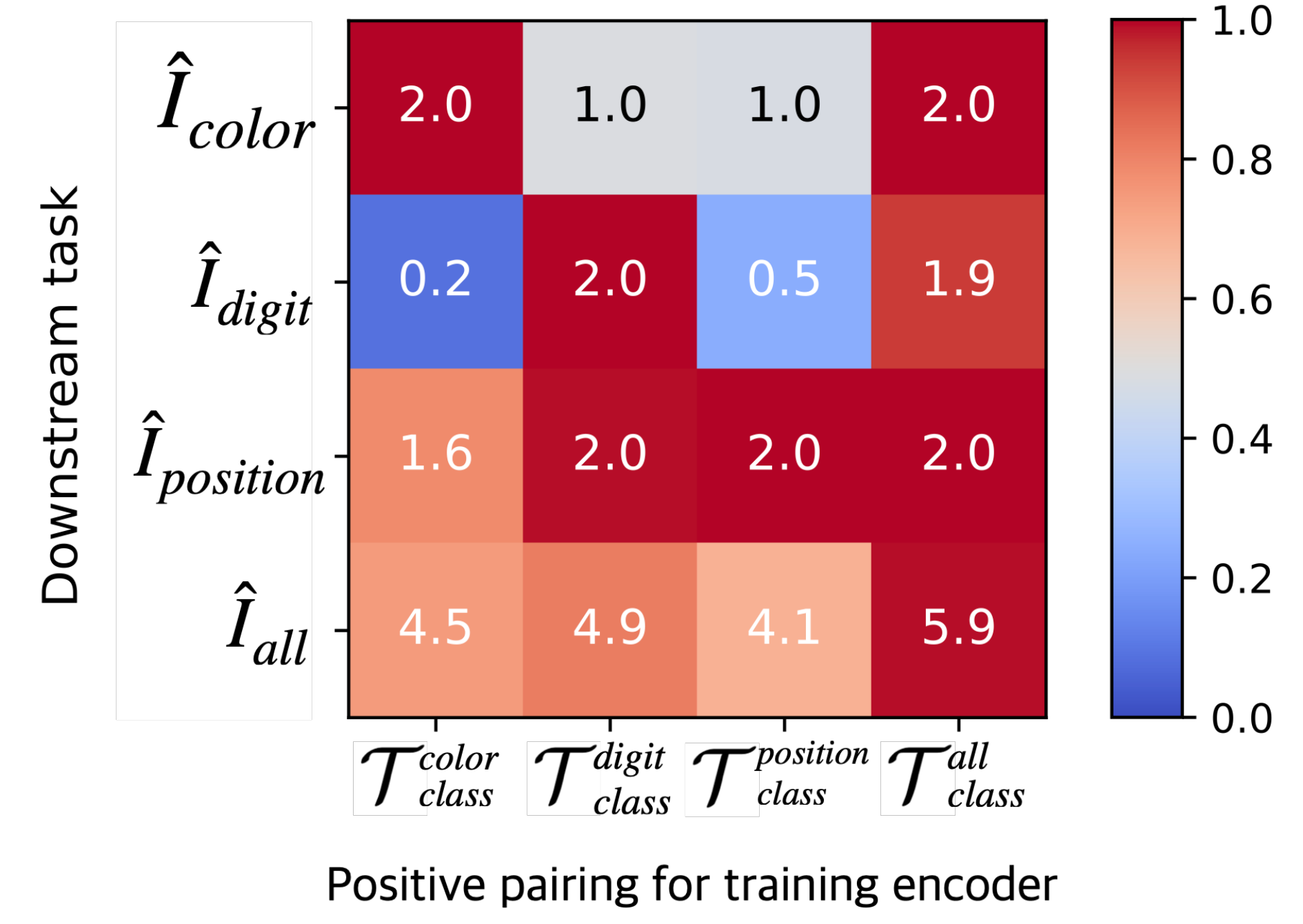}}
    \subfloat[ResNet-50]{
    \includegraphics[width=0.35\linewidth]{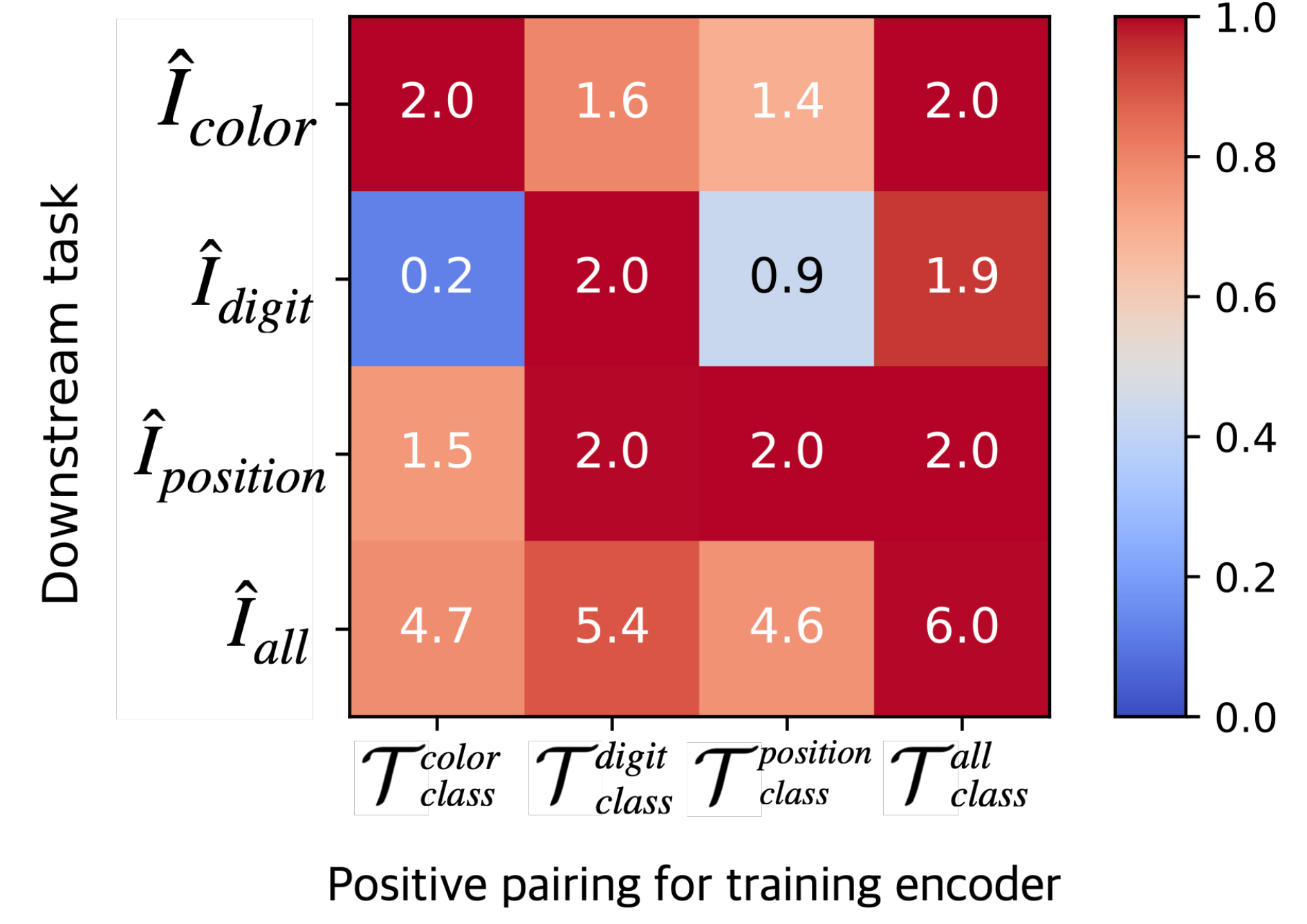}}
    \caption{Post-training MI estimation results for the experiment cases in Figure~\ref{fig:aug.sup}. Strength scale is relative to the $H(C)$, where $H(C)$ is $6.0$ for $\hat{I}_{all}$ and $2.0$ for the others. 
    }
    \label{fig:aug.mi}
\end{figure}

MI estimation results that correspond to the experiment cases of Figure~\ref{fig:aug.sup} are shown in Figure~\ref{fig:aug.mi}.

\section*{C. MI maximization vs. NCE}
\label{supp:subsection_appendix_MI_maximization_vs_NCE}

For the contrastive learning to be equivalent to an MI maximization, the negative term (the denominator in Eq.~\eqref{eq:infoNCE_loss}) normalized by $(2K-1)$ needs to be an asymptotic estimation of the partition function $Z(y)(=\mathbb{E}_{p(y)}[e^{f(x,y)}])$~\citep{poole2019variational}. This requirement can be fulfilled by drawing the negative samples with a uniform distribution over the entire training dataset. In practice, the negative samples in Eq.~\eqref{eq:infoNCE_loss} are chosen as the samples in the mini-batch, primarily for the computational efficiency.

In contrast to the viewpoint of MI maximization, the viewpoint of \textit{Noise Contrastive Estimation}~(NCE) in~\citep{gutmann2010noise} does not require the negative samples to be drawn from the marginal distribution. Instead, the negative samples can be drawn from any reasonable distribution including random noise such as Gaussian noise. Interestingly, both viewpoints were addressed in the original CPC work~\citep{oord2018cpc}, but the relationship between the two viewpoints was not clarified. Here, we provide an experiment to show that the negative samples do not need to be drawn from the marginal distribution. In fact, we can enhance the performance of contrastive learning by carefully manipulating the negative sampling.

Before proceeding, we define four new datasets. CIFAR-5A and CIFAR-5B are disjoint datasets created from CIFAR-10. CIFAR-5A contains all the examples of the first five classes of CIFAR-10 and CIFAR-5B contains all the examples of the last five classes of CIFAR-10. CIFAR-50A and CIFAR-50B are created in a similar way from CIFAR-100 (first fifty classes of CIFAR-100 and last fifty classes of CIFAR-100).
The experimental results are shown in Table~\ref{tab:negative_sampling}.
The positive pairs are always drawn from the original dataset $\mathcal{D}$ (CIFAR-5A or CIFAR-50A), but the negative samples are drawn from the negative sampling dataset $\mathcal{D}^-$. As expected, performance degradation can be observed when $\mathcal{D}^-$ is one of PACS-(cartoon, art, photo, and sketch)~\citep{li2017deeper} or uniform random noise. When $\mathcal{D}^-$ is CIFAR-5B, however, the performance is improved by 1.92\%. The same observations can be made for CIFAR-50A, with the improvement of 0.77\%. The experiment results indicate that we can improve the linear evaluation performance by carefully choosing $\mathcal{D}^-$ for negative sampling.
In our experiments, the performance was enhanced by choosing a dataset whose distribution slightly diverges from the true marginal distribution (CIFAR-5B and CIFAR-50B are not the marginals but at least they come from the same source datasets of CIFAR-10 and CIFAR-100).

\begin{table}[ht!]
    \centering
    \captionsetup{position=top}
    \caption{The effect of negative sampling dataset $\mathcal{D}^-$. Linear evaluation performance can be affected by choosing negative samples from a related or an unrelated dataset. (a) CIFAR-5A: For contrastive learning of CIFAR-5A dataset, the best performance is achieved by choosing the negative samples from CIFAR-5B dataset (i.e., not from CIFAR-5A dataset). (b) CIFAR-50A: For contrastive learning of CIFAR-50A dataset, the best performance is achieved by choosing the negative samples from CIFAR-50B dataset (i.e., not from CIFAR-50A dataset).}
    \label{tab:negative_sampling}
    \subfloat[$\mathcal{D}=$ CIFAR-5A]{
    \resizebox{\textwidth}{!}{%
    \begin{tabular}{lcccccccc}
    \toprule
    $\mathcal{D}^-$ & CIFAR-5A (Baseline: InfoNCE loss)
    & CIFAR-5B & PACS-C & PACS-A & PACS-P & PACS-S & Uniform random \\\midrule
    Accuracy (\%) & 85.70
    & \textbf{87.62} & 83.14 & 81.98 & 81.14 & 80.86 & 79.80 \\\bottomrule
    \end{tabular}
    }} \\
    \subfloat[$\mathcal{D}=$ CIFAR-50A]{
    \resizebox{\textwidth}{!}{%
    \begin{tabular}{lcccccccc}
    \toprule
    $\mathcal{D}^-$ & CIFAR-50A (Baseline: InfoNCE loss)
    & CIFAR-50B & PACS-C & PACS-A & PACS-P & PACS-S & Uniform random  \\\midrule
    Accuracy (\%) & 59.56
    & \textbf{60.34} & 49.52 & 51.16 & 50.44 & 43.40 & 33.92 \\\bottomrule
    \end{tabular}
    }}
\end{table}

\end{document}